\documentclass[final,5p,times,twocolumn,numbers]{elsarticle}
\usepackage[dvipsnames]{xcolor}
\usepackage[T1]{fontenc}
\usepackage[utf8]{inputenc}
\usepackage{amsmath,amssymb,amsfonts,amsthm}
\usepackage{graphicx,booktabs,multirow,array,tabularx}

\usepackage{algorithm,algpseudocode}
\usepackage{hyperref,url,float,subcaption,enumitem}
\hypersetup{hidelinks}
\usepackage{tikz}
\usetikzlibrary{positioning,shapes.geometric,arrows.meta,calc}

\newtheorem{definition}{Definition}
\newtheorem{remark}{Remark}
\newenvironment{bluereview}{%
  \begingroup\color{black}%
}{%
  \endgroup%
}
\newenvironment{greenreview}{%
  \begingroup\color{black}%
}{%
  \endgroup%
}

\newcommand{\Sscore}{\mathcal{S}}
\journal{}
\setlist{nosep,leftmargin=*}

\begin{document}
\begin{frontmatter}
\title{Training-Free Clinical Reasoning through Medical Ontologies and Cognitive Mapping: A Symbolic–Probabilistic Knowledge Graph Framework}
\author[1]{Surajit Das\corref{cor1}}
\cortext[cor1]{Corresponding author: Surajit Das (mr.surajitdas@gmail.com). Code and implementation materials are available upon request.}

\begin{abstract}
\begin{bluereview}
Most clinical prediction systems learn patient-variable--outcome
associations; we investigate an alternative, training-free
diagnostic paradigm that maps patient observations onto explicit
medical knowledge. CKG Reasoner integrates candidate-specific
Evidence Feature Nodes, patient--reference matching, a bounded
Information Gate, knowledge-weighted evidence accumulation,
disease-level similarity, and decisive clinical rules.
Missing-aware normalization and evidence-coverage auditing
distinguish absent from unavailable evidence. Candidate ranking
remains separate from outcome-label-independent K-means
clustering, which uses evidence strength, relative magnitude,
directional similarity, and evidence completeness to derive
cohort-level diagnostic assignments.

Retrospective evaluation across six clinical cohorts comprising
four dengue datasets ($N=1000,1523,989,1018$), one malaria
dataset ($N=2190$), and one influenza dataset ($N=4569$)
yielded positive-class F1 scores under a uniform, label-free,
cohort-fitted $K=2$ partition protocol of
0.996, 0.634, 0.936, and 0.917 for dengue,
0.695 for malaria, and 0.842 for influenza.
Corresponding all-record accuracies were 0.996, 0.558,
0.914, 0.893, 0.707, and 0.906, respectively.
All six cohorts achieved full partition-decision coverage
under a uniform $K=2$ protocol using the previously
frozen package and disease-specific knowledge representations.

Neither the knowledge-grounded scorer nor unsupervised
clustering is fitted using outcome labels. K-means uses only
the four derived evidence coordinates, not raw dataset
predictors or target variables. Logistic
regression provides a supervised baseline. Influenza
results incorporate confirmatory molecular PCR and do
not represent independent pre-test prediction.
The findings characterize knowledge-grounded evidence separation,
auditability, and evidence sensitivity rather than establish
prospective clinical validity or comparative superiority.
FOL/LLM-based clinical explanation remains an
unevaluated extension.
\end{bluereview}
\end{abstract}

\begin{keyword}
training-free clinical reasoning \sep medical ontology \sep knowledge representation \sep cognitive mapping \sep knowledge graphs \sep symbolic--probabilistic inference \sep explainable AI \sep clinical decision support
\end{keyword}
\end{frontmatter}

\section{Introduction}

Disease diagnosis requires the integration of heterogeneous evidence,
including symptoms, serology, laboratory findings, demographics, imaging,
and clinical context. These observations do not have equivalent diagnostic
roles: a finding may provide weak support, strong support, contradiction,
decisive confirmation, or exclusion, while its clinical importance and
disease specificity may also differ. Outcome-fitted classifiers address
this problem by learning empirical feature--outcome associations from
labeled observations, particularly when domain rules are unknown or
difficult to specify explicitly. Such models can provide strong predictive
performance within a given population but may degrade under distribution
shift or confounding when these factors are not explicitly modeled or
controlled. Moreover, a learned mapping need not explicitly preserve
distinctions among evidential support, contradiction, clinical importance,
specificity, decisive findings, and unresolved evidence that are important
for translational healthcare. Knowledge-based systems make disease--finding
relations explicit but may themselves be brittle under incomplete or
ambiguous evidence. Probability/rule hybrids, ontology-supported diagnosis,
and weighted clinical knowledge graphs (KGs) are established
\cite{DEDUP_01384,DEDUP_01222,DEDUP_00249,DEDUP_00010,DEDUP_03285};
the contribution of the present work therefore does not rest on any one of
these ingredients in isolation.

\paragraph{Conceptual premise.}

The present framework investigates an alternative, knowledge-grounded
modeling paradigm. Medical knowledge is first represented explicitly
through ontology-grounded, disease-specific structures; patient observations
are then cognitively mapped onto these representations; and diagnostic
support is obtained through symbolic--probabilistic reasoning over the
resulting patient--knowledge correspondence. ``Training-free'' is used
specifically to mean that patient outcome labels do not train, fit,
fine-tune, calibrate, or optimize the diagnostic reasoner. The method is
therefore more precisely outcome-label-independent, knowledge-engineered,
and parameter-frozen during cohort evaluation; it is not parameter-free or
assumption-free. The package and its disease-specific knowledge representations were developed independently of all six evaluation datasets and frozen before their evaluation; none of those datasets was used to construct or revise the package. Clinical evidential and measurement parameters are mapped
to fixed ordinal scales that computationally encode specified medical
knowledge rather than outcome-learned parameters.

Within this paradigm, CKG Reasoner isolates cross-sectional diagnostic
reasoning from temporal progression and explicitly separates operations
that are often conflated: patient--knowledge correspondence and evidence
accumulation, relative candidate ranking, decisive
pathognomonic/exclusionary rules, evidence-completeness auditing, and
unsupervised partition-based diagnostic assignment from the post-reasoning representation. These
operations answer different questions. Candidate ranking expresses relative
support among represented disease hypotheses and is not a calibrated
posterior probability or a prospectively validated diagnosis. Decisive
clinical rules preserve explicit confirmation or exclusion semantics.
Evidence completeness describes how much diagnostically relevant evidence
was available without treating missingness as contradiction. Unsupervised partitioning uses quantities produced by the fixed reasoner to assign cohort-level diagnostic groups. It therefore participates in the reported classification pathway, although it does not modify candidate evidence accumulation, clinical rules, or ranking.

\begin{bluereview}
The objective of the present study is therefore not to establish predictive
superiority over outcome-fitted classifiers. Supervised models estimate
empirical mappings from patient variables to outcomes using labeled
observations, whereas the present framework investigates whether explicitly
represented medical knowledge can be operationalized as a fixed,
patient-activated reasoning process without fitting the diagnostic mapping
to cohort outcome labels. Predictive discrimination alone would consequently
not constitute a like-for-like evaluation of the principal contribution.
The primary questions in this initial study are whether the encoded
reasoning process is computationally coherent, auditable at the level of
individual evidence contributions, robust to plausible perturbations,
sensitive to clinically relevant evidence removal, and capable of exposing
the provenance and limitations of its conclusions. Outcome-fitted models
remain useful contextual benchmarks, and comparative predictive and
clinical-utility studies remain important subsequent validation questions,
particularly under matched external cohorts, aligned information
availability and clinical tasks, and prospective evaluation of
discrimination, calibration, selective prediction, and clinical utility.
\end{bluereview}

\begin{bluereview}
Unsupervised partitioning constitutes the diagnostic assignment stage used to produce the reported cohort-level classification labels, rather than merely an evaluation-level visualization. The clustering uses only the four derived evidence coordinates; neither raw dataset predictors nor target/outcome labels are clustering inputs. The reported K-means partitioning operates on the post-reasoning representation
generated by the fixed inference process and does not alter candidate
evidence accumulation, decisive clinical rules, or candidate ranking.
Documented outcomes are subsequently used to calculate F1, accuracy, and related metrics for the partition-derived diagnostic assignments. These metrics evaluate the cohort-level assignments produced by the integrated, outcome-label-independent diagnostic-assignment stage. They are not
interpreted as prospective clinical performance, calibrated diagnostic
accuracy, or substitutes for an externally validated clinical decision
rule. Supervised logistic-regression results are supplied in the manuscript as a contextual baseline for all six cohorts; their underlying fitting scripts and split provenance were not independently reproduced from the six CKG notebooks. Its held-out test metrics and the cohort-fitted CKG partition metrics follow different evaluation protocols, so their numerical comparison is descriptive rather than evidence of matched predictive superiority. Random forests, gradient-boosted models, and neural classifiers were not established as executed baselines in the supplied experiments.
\end{bluereview}

\begin{bluereview}
The empirical study evaluates six heterogeneous clinical cohorts using the
same inference machinery with disease-specific knowledge representations:
four dengue cohorts with different evidence profiles, one malaria cohort,
and one influenza cohort. All six cohorts A--F were evaluated using the previously frozen CKG Reasoner package and disease-specific knowledge representations; none was used to develop or modify the package or its knowledge representations. A uniform $K=2$ K-means protocol was applied across all six cohorts; outcome labels were excluded from package construction, schema mapping, knowledge-grounded scoring, and K-means fitting. These infectious-disease datasets provide
clinically overlapping but heterogeneous evidence regimes through which the
behavior of the fixed reasoning architecture can be examined. No cohort
outcome labels are used to train, fit, fine-tune, calibrate, or optimize the
reasoner or its disease-specific knowledge representation. Negative records
are not relabelled as unsupported alternative diseases. Accordingly, the
six cohorts evaluate the architecture across separate disease-specific
binary tasks rather than constituting a validated three-class
dengue--malaria--influenza differential-diagnosis benchmark.
\end{bluereview}

\begin{greenreview}
The principal contributions of this work are:
\begin{enumerate}

\item An outcome-label-independent, candidate-conditioned
symbolic--probabilistic reasoning architecture integrating
medical ontologies, cognitive patient--knowledge mapping,
evidence accumulation, disease ranking, decisive clinical
rules, and unsupervised diagnostic assignment.

\item Disease-specific Evidence Feature Nodes (EFNs)
that preserve canonical clinical states and explicitly
distinguish diagnostic importance, specificity,
patient--reference correspondence, positive support,
contradiction, and pathognomonic/exclusionary roles.

\item A knowledge-engineered inference procedure with
frozen diagnostic parameters, missing-aware evidence
normalization, and separate evidence-completeness
auditing. The latter identifies unresolved critical
evidence and quantifies Confidence Score (CS)
without incorporating it into candidate ranking.

\item An extensible, inference-faithful architecture
providing patient-specific reasoning traces and
evidence provenance, with explicit design pathways
toward uncertainty-region identification, interactive
evidence acquisition, FOL-compatible explanations,
constrained language generation, and real-time
clinical decision support. These extensions are
distinguished from experimentally validated capabilities.

\item Retrospective evaluation across six heterogeneous
dengue, malaria, and influenza cohorts using frozen
knowledge representations and a uniform,
outcome-label-independent clustering protocol,
supported by available component-ablation,
perturbation, evidence-withholding,
parameter-sensitivity, and initialization-stability
analyses.

\end{enumerate}
\end{greenreview}

The separate binary cohorts are not presented as a three-class
differential-diagnosis benchmark, and the study does not claim predictive
superiority over outcome-fitted classifiers.

\paragraph{Novelty statement.}
\begin{bluereview}
The novelty of CKG Reasoner lies in the design and integration
of an outcome-label-independent, ontology-grounded,
candidate-conditioned symbolic--probabilistic reasoning
architecture, rather than in its individual components.
Through cognitive patient--knowledge mapping, the framework
operationally separates clinical importance, diagnostic
specificity, patient--reference similarity, positive and
contradictory evidence, decisive clinical rules, and
evidence completeness. This semantic decomposition enables
patient-specific disease ranking, auditable reasoning,
explicit evidence provenance, and a separate unsupervised
diagnostic-assignment pathway without fitting diagnostic
parameters to patient outcome labels.

An additional architectural contribution is its extensibility
toward translational healthcare. Unlike approaches restricted
to fixed-output classification, the design preserves the
information necessary to investigate explicit uncertainty
regions, identify unresolved critical evidence, request
additional patient information, and revise diagnostic
conclusions through further reasoning. Its inference-faithful
traces also provide a foundation for clinically grounded
explanations and prospective real-time, patient-wise
clinical decision support. These research opportunities
arise from the architecture's explicit evidence semantics,
modular reasoning stages, and separation of ranking,
decision-making, and evidence completeness.

The contribution is therefore both an implemented,
auditable reasoning methodology and an extensible
architecture for future interactive and explainable
clinical AI, complementary to outcome-fitted classifiers.
Real-time deployment, interactive evidence acquisition,
explicit uncertainty-region validation, and clinical
validation of FOL/LLM explanations remain future work.
\end{bluereview}

\section{Related Work and Positioning}

\subsection{From expert rules to probabilistic and ontology-based diagnosis}
Probability--rule hybrids are longstanding: CLAUDE combined rule and probabilistic experts through a neural reconciler \cite{DEDUP_01384}, while probabilistic induction learned weighted medical rules from historical cases \cite{DEDUP_01400}. Ontology-based systems later combined probabilistic inference \cite{DEDUP_01222}, fuzzy rules, semantic similarity and hierarchical weighting \cite{DEDUP_00138,DEDUP_00162}, or symptom-dependency-aware Naive Bayes \cite{DEDUP_00249}. Thus neither probability-plus-rules nor semantic uncertainty is novel here; the distinction is the graph-structured separation of evidence semantics and decisive clinical roles.

Bayesian models likewise provide interpretable differential reasoning: expert-knowledge BNs \cite{SB_DEDUP_00528}, a prospectively designed anterior-uveitis BN \cite{SB_DEDUP_00404}, Siamese BNs emphasizing negative evidence \cite{SB_DEDUP_00169}, dengue BNs with/without NS1 \cite{SB_DEDUP_00057}, and large leaky noisy-OR networks derived from Orphanet/HPO \cite{DEDUP_00352}. These establish uncertainty and negative evidence as prior art; the present question is their integration with richer graph semantics and hard clinical overrides without outcome-driven fitting.

\subsection{Knowledge graphs for diagnostic reasoning}
Knowledge-graph CDSS research spans graph representation and reasoning \cite{SB_DEDUP_00105}, learned interactive policies such as DKDR and RDKG \cite{DEDUP_01201,DEDUP_01193}, weighted symptom--syndrome paths \cite{SB_DEDUP_00479}, embedding-refined diagnosis paths \cite{DEDUP_00180}, and recent sleep-disorder reasoning \cite{DEDUP_02281}. A particularly close headache engine uses ICHD-3, weighted graph matching, fulfilled-criteria weights, and exclusion penalties \cite{DEDUP_03285}; therefore transparent exclusion-aware graph scoring is not claimed as unique.

Graph uncertainty is also established. A semantic clinical KG uses likelihood-ratio weights and patient-specific conditional edges for hypothesis ranking \cite{DEDUP_00010}, while uncertain KGs have been constructed from personal EHRs \cite{DEDUP_01189}. The present framework differs by preserving canonical disease features, clinical importance, specificity, patient/reference agreement, support, contradiction, and hard pathognomonic/exclusionary roles as distinct inference quantities rather than primarily weighted relations.

\subsection{EHR-centered, patient-specific, and learned clinical graphs}
EHR-oriented KGs address fragmented-data integration and CDSS workflow \cite{DEDUP_00212,DEDUP_01170,SC_01}, whereas this work assumes a pre-specified clinical knowledge model and studies inference semantics. Modern patient-specific systems include D$^{2}$KGMed, which uses LLM-guided graph construction and supervised fine-tuning \cite{DEDUP_00290}; DR.KNOWS, which ranks UMLS paths with graph/LLM components \cite{SC_03}; and KGDAgents \cite{DEDUP_01192}. Their value supports patient-specific structured reasoning, but their diagnostic behavior depends on learned, graph-neural, agentic, or language-model components; here patient records activate a fixed knowledge-driven procedure whose parameters are not estimated from outcome labels.

\subsection{Neuro-symbolic and LLM-guided reasoning}
Neuro-symbolic CDSSs combine deep learning with symbolic reasoning \cite{DEDUP_00001,DEDUP_00227}, including rule engines coupled to neural language processing \cite{DEDUP_00749}. Structured-knowledge LLM systems include ReCLLaMA \cite{DEDUP_01211}, DR.KNOWS \cite{SC_03}, and epistemologically guided diagnostic reasoning \cite{DEDUP_00114}. These improve structure and reviewability but retain learned/generative components. Here the implemented numerical evidence and rule traces provide a foundation for clinically grounded explanation; FOL serialization and LLM-based verbalization are prospective extensions rather than evaluated contributions in this study.

\subsection{Closest methodological gap}

The literature does \emph{not} support broad novelty claims for knowledge
graphs, probabilistic reasoning, clinical rules, negative evidence, semantic
similarity, patient-specific reasoning, or explainability, each of which has
clear precedent
\cite{DEDUP_01384,DEDUP_00138,SB_DEDUP_00169,DEDUP_00010,DEDUP_03285}.
The methodological gap concerns how these elements are organized
within the diagnostic inference process. Relatively few systems preserve
clinical importance, diagnostic specificity, patient--reference agreement,
positive support, contradiction, and decisive clinical roles as distinct
operational quantities within a single outcome-label-independent,
patient-activated reasoning architecture, while also allowing
pathognomonic/exclusionary findings to remain operationally distinct from
graded evidence and exposing a patient-specific inference trace constructed
from the same quantities that generate the diagnostic result. Against the
representative systems summarized in Table~\ref{tab:relatedcomparison}, the
contribution of the present work is therefore best characterized by this
explicit semantic decomposition, its integration within a fixed,
auditable, candidate-conditioned symbolic--probabilistic reasoning
architecture, and inference-faithful patient-level traceability, rather than
by any individual component in isolation.

\begin{table*}[t]
\centering
\caption{Representative diagnostic reasoning systems and their relationship to the proposed framework. ``Learned'' indicates that a substantive diagnostic component is estimated or fine-tuned from data.}
\label{tab:relatedcomparison}
\scriptsize
\setlength{\tabcolsep}{3.5pt}
\renewcommand{\arraystretch}{1.12}
\begin{tabularx}{\textwidth}{p{0.105\textwidth}p{0.19\textwidth}ccccccX}
\toprule
\textbf{Study} & \textbf{Primary representation} &
\textbf{Prob.} & \textbf{Rules} & \textbf{Neg. ev.} &
\textbf{Patient-specific} & \textbf{Learned} & \textbf{Intrinsic trace} &
\textbf{Main distinction from this work}\\
\midrule
\cite{DEDUP_01384} & Hybrid rule/probabilistic experts & Yes & Yes & -- & Yes & Yes & Partial & Combines expert outputs through a neural network; no explicit clinical KG evidence semantics.\\
\cite{DEDUP_00138} & Ontology + fuzzy rule system & Fuzzy & Yes & Partial & Yes & Partial & Yes & Semantic similarity and fuzzy inference, but no separate support/contradiction channels or decisive rule override layer.\\
\cite{SB_DEDUP_00169} & Siamese Bayesian networks & Yes & -- & Yes & Yes & Yes & Yes & Explicitly addresses negative evidence, but within a learned BN formulation rather than a disease-specific EFN graph.\\
\cite{DEDUP_01201} & KG + deep reinforcement learning & -- & -- & -- & Yes & Yes & Partial & Learns interactive diagnostic policy over the KG.\\
\cite{SB_DEDUP_00479} & KG reasoning paths + dynamic weights & Yes & -- & Partial & Yes & Partial & Yes & Weighted paths for TCM syndrome reasoning; no hard pathognomonic/exclusionary override semantics.\\
\cite{DEDUP_00010} & Semantic KG with LR-weighted/conditional edges & Yes & Conditional & Yes & Yes & No/limited & Yes & Very close weighted transparent reasoning; evidence remains primarily edge-weight based rather than explicitly separated clinical attributes.\\
\cite{DEDUP_03285} & ICHD-3 KG + weighted matching & Weighted & Criteria & Yes & Yes & No/limited & Yes & Transparent differential engine with exclusion penalties; no probabilistic evidence/similarity/hard-rule decomposition used here.\\
\cite{DEDUP_00290} & Patient-specific dynamic diagnostic KG + LLM & -- & -- & -- & Yes & Yes & Yes & Dynamic patient graph is learned/refined through LLM and supervised fine-tuning.\\
\cite{SC_03} & UMLS KG paths + graph model + LLM & -- & -- & -- & Yes & Yes & Yes & Retrieves patient-specific reasoning paths, but prediction depends on learned graph/LLM components.\\
\cite{DEDUP_01211} & Neuro-symbolic LLM agent + KG & -- & Yes & Partial & Yes & Yes & Yes & Structured agentic reasoning over free text; generative/learned inference remains central.\\
\textbf{Proposed} & Disease-specific EFNs + patient evidence graph & Yes & Yes & Yes & Yes & \textbf{No outcome fitting} & Yes & Separates importance, specificity, similarity, support, contradiction, and decisive clinical roles within one fixed inference procedure.\\
\bottomrule
\end{tabularx}
\end{table*}

\section{Methodology}
\label{sec:methodology}
\subsection{Clinical Knowledge Representation}
\subsubsection{Reference knowledge graph}
Let $\mathcal{G}_R=(V_R,E_R)$ denote a reference medical knowledge graph constructed from clinical guidelines, pathological observations, laboratory biomarkers, radiological findings, epidemiological evidence, and expert knowledge. Nodes represent symptoms, laboratory measurements, biomarkers, imaging findings, contextual variables, intermediate clinical concepts, or diseases. The graph serves as a cross-sectional evidence structure linking observed patient features to candidate disease hypotheses without implying temporal progression.

The representation separates two complementary aspects of diagnostic
reasoning: (i) the \emph{patient--reference aspect}, describing the
correspondence between an individual's observed features and the
candidate-specific canonical representation; and (ii) the
\emph{disease-knowledge aspect}, describing the diagnostic meaning of each feature for a candidate disease. This semantic separation between patient–reference evidence and disease-specific knowledge is essential because the same observed feature may have different diagnostic significance across competing diseases

\paragraph{Patient--reference aspect}

For feature $F_i$ under candidate disease $D_j$, the patient--reference aspect
compares the activated patient representation of that feature with its
candidate-specific canonical representation. A feature may contain several
clinically distinct components; for example, a fever feature may contain
onset, duration, and temperature components. The patient observation itself
is unchanged across candidates, but its activation is evaluated relative to
the canonical state specified by each disease model. The mathematical
construction of component activation, feature-level directional agreement,
relative magnitude, and the resulting bounded Information Gate is given in
Section~\ref{sec:operational} after component activation. This
patient--reference correspondence is distinct from fixed disease-knowledge
attributes such as diagnostic role, clinical importance, support strength,
and contradiction magnitude.

\paragraph{Disease-specific knowledge attributes and Evidence Feature Nodes}
\label{sec:unified_attributes}

For feature $i$ under candidate disease $D_j$, $g_{ij}$ is the diagnostic-role
weight/direction, $I_{ij}$ clinical importance, $S_{ij}$ support strength,
and $C_{ij}$ contradiction magnitude. Patient--reference correspondence is
computed separately from activated feature components through the Information
Gate $IG_{ij}$ in Section~\ref{sec:operational}. The fixed disease-knowledge
semantics and accompanying encodings are as follows:

\subparagraph{Clinical importance.}
Clinical importance $I_{ij}$ denotes the \emph{practical diagnostic importance} of feature $F_i$ for candidate disease $D_j$, distinct from disease specificity. Features with the same diagnostic role may differ in importance, and a highly disease-specific feature need not be maximally important, because practical contribution may depend on availability, measurement reliability, disease stage, and clinical context. It is defined as
\[
I_{ij}\in\{1,\,0.75,\,0.50,\,0.25\},
\]
corresponding to \emph{Critical}, \emph{Major}, \emph{Moderate}, and \emph{Minor}, respectively.

\subparagraph{Diagnostic role and disease specificity.}
The diagnostic role $g_{ij}$ denotes the \emph{disease specificity and diagnostic direction} of feature $F_i$ for candidate $D_j$, i.e., how characteristic or diagnostically informative the feature is for that disease. It is candidate-specific, so the same feature may have different roles across diseases. Unlike patient--disease similarity, which measures correspondence between a patient's observed feature representation and the candidate-specific canonical disease representation, specificity is encoded medical knowledge about the feature--disease relationship. Thus, a highly specific feature may be absent or poorly matched in a patient, while a patient may strongly match a nonspecific feature shared across diseases. The scale is
\[
g_{ij}\in\{1,\,0.85,\,0.70,\,0.50,\,0.30,\,0.15,\,-0.50\},
\]
corresponding to \emph{Pathognomonic}, \emph{Hallmark}, \emph{Major}, \emph{Supportive}, \emph{Associated}, \emph{Nonspecific}, and \emph{Exclusionary}, respectively.

\subparagraph{Support strength.}
Support strength $S_{ij}$ denotes the \emph{positive evidential strength} contributed by feature $F_i$ for $D_j$ when the patient observation is present and appropriately matched, distinct from specificity and clinical importance. Hence, features with identical roles and importance may differ in support: two \emph{Hallmark} features may provide strong versus moderate support if one is less consistent, more context-dependent, or more susceptible to alternative explanations. Conversely, an uncommon feature may have moderate practical importance yet provide decisive support when present if strongly associated with $D_j$. Thus, $S_{ij}$ independently represents the evidential consequence of an observed match:
\[
S_{ij}\in\{1,\,0.8,\,0.6,\,0.3,\,0\},
\]
corresponding to \emph{Decisive}, \emph{Strong}, \emph{Moderate}, \emph{Slight}, and \emph{None}, respectively.

\subparagraph{Contradiction magnitude.}
Contradiction magnitude $C_{ij}$ denotes the \emph{negative evidential strength} against $D_j$ when the observed state of $F_i$ conflicts with its candidate-specific expectation. It is not simply the inverse of $S_{ij}$ because agreement and disagreement may have asymmetric consequences. For example, a highly characteristic but uncommon manifestation may provide strong or decisive support when present ($S_{ij}=0.8$ or $1$) but little contradiction when absent ($C_{ij}=0.2$). Conversely, a finding expected in nearly all affected patients but common in other diseases may provide only moderate support when present ($S_{ij}=0.6$) yet strong contradiction when unexpectedly absent ($C_{ij}=0.7$). Thus, positive and negative evidence are represented independently:
\[
C_{ij}=\left|\min\!\left(0,\mathrm{Contra}_{ij}\right)\right|
\in\{1,\,0.7,\,0.4,\,0.2,\,0\},
\]
corresponding to \emph{Decisive}, \emph{Strong}, \emph{Moderate}, \emph{Mild}, and \emph{None}, respectively.

Thus $C_{ij}$ is the nonnegative magnitude used by the negative-evidence equation.

\begin{definition}[Evidence Feature Node]
\[
\mathrm{EFN}_{ij}=(F_i,g_{ij},I_{ij},S_{ij},C_{ij},T_i),
\]
where $F_i$ identifies the feature and $T_i$ is optional non-temporal metadata.
The candidate-specific canonical feature state is stored with the reference
knowledge representation. Patient observations are not part of the fixed EFN;
they are activated against that canonical state during inference to obtain
$IG_{ij}$.
\end{definition}

\begin{remark}[Candidate dependence of an EFN]
An EFN is candidate-conditioned rather than a generic clinical-feature node. The same feature $F_i$ may occur across candidate disease graphs with different canonical states, diagnostic roles, importance values, support strengths, and contradiction magnitudes. The patient observation itself remains unchanged; only the disease-specific reference and its diagnostic interpretation vary across candidates.
\end{remark}

\subparagraph{Outcome-label-independent evaluation.}
The inference machinery and disease-specific knowledge
representations were developed independently of all six
evaluation cohorts and frozen before evaluation.
Dataset-specific schema mapping preserved the underlying
knowledge. All cohorts underwent uniform, label-free
K-means partitioning ($K=2$) using four-dimensional
post-reasoning vectors.

Classification metrics were derived from partition
assignments rather than candidate-ranking scores.
K-means uses only the four derived, knowledge-grounded
coordinates, not raw dataset predictors or target labels.
Reference outcomes are used only after assignment to
calculate evaluation metrics.
The supervised logistic-regression baseline follows
a different evaluation protocol; comparisons are
therefore descriptive rather than evidence of
matched predictive superiority.

\subsection{Hybrid Diagnostic Reasoning}

\subsubsection{Training-free design}
\label{sec:trainingfree}

\begin{bluereview}
\subparagraph{Frozen experimental specification.}

The medical ontology and disease-specific knowledge graphs,
including canonical clinical features, diagnostic roles,
clinical-importance weights, support and contradiction
strengths, and hard-rule semantics, were constructed
principally from established WHO, CDC, and PAHO clinical
guidance and medical knowledge. The corresponding numerical
scales, scoring coefficients, activation functions, and
decision thresholds are explicitly defined knowledge-engineering
choices rather than parameters learned from patient outcomes
or necessarily prescribed by those guidelines.

The ontology, disease-specific knowledge representations,
and inference parameters were developed independently of
all six evaluation datasets and frozen before evaluation.
Patient records supply observations for patient--knowledge
mapping; reference outcomes are reserved exclusively for
post-assignment evaluation. No patient outcome labels are
used to train, fit, fine-tune, calibrate, or optimize the
diagnostic reasoner or its knowledge representations.

Canonical \texttt{default} components are handled through
explicit patient--reference matching. Repeated feature codes
occurring at distinct disease stages retain stage-specific
identities and remain independent evidence nodes.

The scorer preserves $P_{\mathrm{evidence}}$, $r$, and $c$
as disease-level reasoning quantities and separately exports
the outcome-label-independent Confidence Score
(CS; \texttt{Confidence\_Score}), an evidence-completeness
quantity. CS does not enter candidate ranking.

The integrated diagnostic-assignment stage applies a uniform
$K=2$ K-means protocol across all six evaluation cohorts
using the standardized four-dimensional post-reasoning
representation
\[
S_0(D_j)=
[P_{\mathrm{evidence}}(D_j),r_j,c_j,CS_j]^\top.
\]
Candidate ranking uses a separate canonically normalized
additive score defined below. K-means operates exclusively
on the four derived knowledge-grounded evidence coordinates,
not on raw clinical predictors or outcome labels. Its
centroids are fitted independently within each cohort,
making the retrospective assignment procedure transductive
rather than an independent out-of-sample prediction test.
Ground-truth labels enter only after partitioning for
retrospective metric calculation.
\end{bluereview}

\subsubsection{Ontology-grounded cognitive mapping}
\label{sec:cognitive_mapping}
\begin{bluereview}
The central operation of CKG Reasoner is a patient-to-knowledge mapping rather than a learned feature-to-label mapping. Let the medical ontology/knowledge representation define, for candidate disease $D_j$, a set of Evidence Feature Nodes with canonical component states and fixed clinical semantics. For an observed patient $x$, the cognitive map
\[
\mathcal M_j:x\longmapsto
\left\{\mathbf q_{ij},\mathbf r_{ij},\cos\theta_{ij},\rho_{ij},IG_{ij},
g_{ij},I_{ij},S_{ij},C_{ij}\right\}_{i}
\]
constructs a candidate-conditioned representation of how the patient's evidence corresponds to the encoded disease model. Here $\mathbf q_{ij}$ is the activated patient representation, $\mathbf r_{ij}$ is its candidate-specific canonical reference, $\cos\theta_{ij}$ represents directional agreement, $\rho_{ij}$ represents relative magnitude, and $IG_{ij}$ is their bounded feature-level Information Gate. The remaining quantities are disease-knowledge attributes supplied by the ontology rather than learned from the patient cohort.

This use of ``cognitive mapping'' denotes an explicit computational correspondence between observed clinical evidence and structured medical knowledge. It does not imply a model of human cognition. The mapping is candidate-specific: the same observation can have different diagnostic meaning under different disease hypotheses because canonical states, roles, support, contradiction, and decisive-rule semantics belong to the disease representation. The subsequent reasoning layer operates on this mapped representation and never estimates its parameters from cohort outcome labels.

This decomposition is important for auditability. An erroneous decision can, in principle, be localized to the encoded medical knowledge, the patient-to-reference activation/boundary function, the interaction of evidence, a hard rule, or the final aggregation mechanism rather than being hidden inside learned model weights. Conversely, being training-free does not guarantee clinical correctness: the framework can only reason from the quality and contextual validity of the knowledge and mapping functions supplied to it.
\end{bluereview}

\subsubsection{Operational scoring}

\label{sec:operational}
\subparagraph{Component activation.}
Each available patient component $x$ is transformed according to the type of its candidate-specific canonical reference. Let $\kappa_i(x)$ denote the fixed codebook value for a categorical state, $\varepsilon=0.01$ the numerical matching tolerance, $A_i$ a finite admissible set, $[a_i,b_i]$ a canonical interval, and $G_i(x)$ the fixed feature-specific fall-off outside that interval. The component activation is
\begin{equation}
q_i(x)=
\begin{cases}
\kappa_i(x), & \text{binary/categorical},\\
\mathbf 1\{|x-a_i|<\varepsilon\}, & \text{single numerical value},\\
\mathbf 1\{\exists a\in A_i:\ |x-a|<\varepsilon\}, & \text{finite numerical set},\\
\mathbf 1\{x\in A_i\}, & \text{finite categorical set},\\
1, & \text{interval and }x\in[a_i,b_i],\\
G_i(x), & \text{interval and }x\notin[a_i,b_i].
\end{cases}
\label{eq:component_activation}
\end{equation}
For a fixed normal range $[L_i,U_i]$, directional trend references are
\begin{align}
q_i^\downarrow(x)&=
\begin{cases}
0,&x\ge L_i,\\
1-\exp[-5(L_i-x)/L_i],&x<L_i,
\end{cases}\nonumber\\
q_i^\uparrow(x)&=
\begin{cases}
0,&x\le U_i,\\
1-\exp[-4(x-U_i)/U_i],&x>U_i,
\end{cases}
\label{eq:trend_activation}
\end{align}
with values clipped to $[0,1]$. A Normal/No-change reference has unit activation inside $[L_i,U_i]$ and the fixed implementation-specific fall-off outside it. Equations~\eqref{eq:component_activation}--\eqref{eq:trend_activation} therefore define the activation rule for all supported reference types.

Because the canonical reference is candidate-specific, applying these transformations to component $c$ of feature $F_i$ under candidate $D_j$ produces the retained activation $q_{ijc}$. Thus, the same observed patient value can produce different activations under different candidate diseases when their canonical reference states differ. These transformations are bounded knowledge-engineering compatibility scores, not calibrated probabilities, disease prevalence, diagnostic specificity, clinical importance, support strength, or contradiction strength.

\begin{bluereview}
\subparagraph{Feature-level patient--reference correspondence.}
Let feature $F_i$ contain $m_i$ mutually orthogonal component axes
$\mathbf u_{i1},\ldots,\mathbf u_{im_i}$, with
$\mathbf u_{ic}^{\top}\mathbf u_{id}=0$ for $c\neq d$ and
$\|\mathbf u_{ic}\|_2=1$. After the candidate-conditioned activation
procedure above, the patient representation for $F_i$ under disease $D_j$ is
\begin{equation}
\mathbf q_{ij}
=
\sum_{c=1}^{m_i}q_{ijc}\mathbf u_{ic},
\qquad 0\le q_{ijc}\le1.
\label{eq:patient_feature_vector}
\end{equation}
Under the adopted canonical representation, every canonical component is an
orthogonal unit component. Thus
\begin{equation}
\mathbf r_{ij}
=
\sum_{c=1}^{m_i}r_{ijc}\mathbf u_{ic},
\qquad r_{ijc}=1,
\qquad
\|\mathbf r_{ij}\|_2=\sqrt{m_i}.
\label{eq:canonical_feature_vector}
\end{equation}
The canonical disease representation is the reference itself and is not passed
through the patient activation transformation. The component-wise normalized
patient magnitude is therefore
\begin{equation}
p_{ij}
=
\left(
\sum_{c=1}^{m_i}
\left(\frac{q_{ijc}}{r_{ijc}}\right)^2
\right)^{1/2}
=
\left(\sum_{c=1}^{m_i}q_{ijc}^{2}\right)^{1/2},
\label{eq:pij_implementation}
\end{equation}
because $r_{ijc}=1$ for every canonical component. The norm is taken only
across components of the same feature for the same patient and never across
patients. Although each $q_{ijc}$ is bounded by one, $p_{ij}$ is a geometric
magnitude and may exceed one when several components are active; it is not a
probability.

Directional patient--canonical agreement for feature $F_i$ is
\begin{equation}
\cos\theta_{ij}
=
\frac{\mathbf q_{ij}^{\top}\mathbf r_{ij}}
{\|\mathbf q_{ij}\|_2\,\|\mathbf r_{ij}\|_2}
=
\frac{\sum_{c=1}^{m_i}q_{ijc}}
{\sqrt{\sum_{c=1}^{m_i}q_{ijc}^{2}}\sqrt{m_i}},
\label{eq:feature_cosine}
\end{equation}
with $\cos\theta_{ij}=0$ when $\|\mathbf q_{ij}\|_2=0$. Relative magnitude is
\begin{equation}
\rho_{ij}
=
\frac{\|\mathbf q_{ij}\|_2}{\|\mathbf r_{ij}\|_2}
=
\frac{\sqrt{\sum_{c=1}^{m_i}q_{ijc}^{2}}}{\sqrt{m_i}}.
\label{eq:feature_rho}
\end{equation}
Since $0\le q_{ijc}\le1$, both $\cos\theta_{ij}$ and $\rho_{ij}$ lie in
$[0,1]$ for an observed/evaluable feature.

The feature-level Information Gate combines these two complementary forms of
agreement. Let
\begin{equation}
\mathbf z_{ij}
=
\begin{bmatrix}
\cos\theta_{ij}\\
\rho_{ij}
\end{bmatrix},
\qquad
\mathbf z^{*}
=
\begin{bmatrix}
1\\1
\end{bmatrix}.
\end{equation}
The ideal self-reference has $\cos\theta=1$ and $\rho=1$, so its squared
Euclidean magnitude is $2$ and its $L_2$ norm is $\sqrt{2}$. Consequently,
\begin{equation}
IG_{ij}
=
\frac{\|\mathbf z_{ij}\|_2}{\|\mathbf z^{*}\|_2}
=
\sqrt{\frac{\cos^2\theta_{ij}+\rho_{ij}^{2}}{2}},
\qquad 0\le IG_{ij}\le1.
\label{eq:information_gate}
\end{equation}
Thus $IG_{ij}=1$ denotes perfect patient--canonical correspondence for that
feature. Missing or unavailable evidence is handled separately and is not
represented by setting $IG_{ij}=0$.
\end{bluereview}

\begin{bluereview}
\subparagraph{Evidence accumulation.}
Evidence is accumulated separately for each candidate disease. Let $\mathcal O_j$ denote the observed/evaluable features for candidate $D_j$; missing or unavailable features are excluded rather than interpreted as observed absence. For a bounded correspondence value $x\in[0,1]$, define the nonlinear evidence gate
\begin{equation}
f(x;\alpha,\beta,\gamma)
=
\frac{1}{1+\alpha\exp[-\beta(x-\gamma)]},
\label{eq:nonlinear_gate}
\end{equation}
where $\alpha>0$ controls the scale of the exponential term, $\beta>0$ controls transition steepness, and $\gamma$ determines its location. The experiments reported here use
\[
\alpha=0.05,\qquad \beta=10,\qquad \gamma=0.85.
\]
The same gate is applied to positive correspondence and to the complementary mismatch proxy. Feature-level contributions are
\begin{align}
\texttt{pcon}_{ij}&=f(IG_{ij};\alpha,\beta,\gamma)\max(g_{ij},0)I_{ij}S_{ij},\\
\texttt{ncon}_{ij}&=f(1-IG_{ij};\alpha,\beta,\gamma)C_{ij}.
\end{align}
The second expression is a transformed complementary-mismatch proxy, not an independently measured contradictory-reference similarity. Diagnostic role $g_{ij}$ and clinical importance $I_{ij}$ retain their encoded values; no exponential transformation of either attribute is used. Candidate-level evidence totals are
\begin{align}
E^+(D_j)
&=\sum_{i\in\mathcal O_j}\texttt{pcon}_{ij},\\
E^-(D_j)
&=\sum_{i\in\mathcal O_j}\texttt{ncon}_{ij},\\
E_{\rm net}(D_j)
&=E^+(D_j)-E^-(D_j).
\end{align}
No additional multiplicative discount is applied to aggregate negative evidence. Evidence from different candidate diseases is never pooled into one $E_{\rm net}(D_j)$.
\end{bluereview}

\begin{bluereview}
\subparagraph{Missing-aware canonical evidence normalization.}
The implementation does not apply a sigmoid to net evidence. For candidate $D_j$, the canonical denominator is constructed over the same observed/evaluable feature support $\mathcal O_j$ used by the patient numerator. The canonical self-reference sets $IG=1$ for those features and applies the same positive- and negative-evidence equations:
\begin{align}
E_{\mathrm{can}}(D_j;\mathcal O_j)
&=\sum_{i\in\mathcal O_j}
 f(1;\alpha,\beta,\gamma)\max(g_{ij},0)I_{ij}S_{ij}\nonumber\\
&\quad-
\sum_{i\in\mathcal O_j} f(0;\alpha,\beta,\gamma)C_{ij}.
\end{align}
The evidence quantity used downstream is the direct canonical normalization
\begin{equation}
P_{\mathrm{evidence}}(D_j)
=
\frac{E_{\mathrm{net}}(D_j)}{E_{\mathrm{can}}(D_j;\mathcal O_j)},
\label{eq:canonical_normalization}
\end{equation}
when $E_{\mathrm{can}}(D_j;\mathcal O_j)>0$, and $0$ otherwise. Unavailable features therefore neither contribute negative evidence nor enlarge the canonical denominator. This missing-aware construction keeps the numerator and denominator on the same observed support and prevents unequal unobserved feature coverage across candidate disease graphs from mechanically changing the normalized evidence scale. No sigmoid or clipping is applied at this stage; consequently $P_{\mathrm{evidence}}$ is a normalized evidence score rather than a calibrated probability and can, in principle, be negative or exceed one.
\end{bluereview}

\begin{bluereview}
\subparagraph{Diagnostic evidence coverage and confidence annotation.}
Missing-aware normalization answers how strongly the \emph{available} evidence corresponds to a candidate, but it does not by itself quantify how much diagnostically important evidence was available. Two patients can therefore obtain similar $P_{\mathrm{evidence}}$ or $R(D_j)$ values despite substantially different amounts of observed evidence. To expose this distinction without treating missingness as negative evidence, the implementation computes a diagnostic evidence-coverage quantity separate from candidate scoring but included in the default clustering representation.

For candidate $D_j$, let $\mathcal K_j^{\mathrm{eval}}$ denote EFNs that are evaluable in the cohort, i.e., represented by the cohort schema and observed for at least one record. This avoids mechanically penalizing every patient for ontology concepts that the dataset never collects. For EFN $i$, let $m_{ij}^{\mathrm{obs}}$ and $m_{ij}^{\mathrm{exp}}$ denote the numbers of observed and reference-defined components, respectively, and define component availability
\begin{equation}
a_{ij}=\frac{m_{ij}^{\mathrm{obs}}}{m_{ij}^{\mathrm{exp}}},\qquad 0\le a_{ij}\le1,
\label{eq:component_availability}
\end{equation}
with the feature-level observed state used for single/default-component EFNs. Thus $a_{ij}=1$ denotes fully observed, $a_{ij}=0$ fully missing, and intermediate values partially observed evidence.

For non-exclusionary EFNs, canonical diagnostic capacity uses the same positive-evidence semantics as the reasoner,
\begin{equation}
w_{ij}=f(1;\alpha,\beta,\gamma)\max(g_{ij},0)I_{ij}S_{ij},\qquad g_{ij}\ge0.
\label{eq:coverage_weight_positive}
\end{equation}
Because an Exclusionary EFN has no positive-evidence capacity but may be decisively informative, its coverage weight is retained through
\begin{equation}
w_{ij}=f(1;\alpha,\beta,\gamma)|g_{ij}|\max(C_{ij},I_{ij}S_{ij}),\qquad g_{ij}<0.
\label{eq:coverage_weight_exclusionary}
\end{equation}
Diagnostic Evidence Coverage (DEC) is then
\begin{equation}
\mathrm{DEC}_j(x)=
\frac{\sum_{i\in\mathcal K_j^{\mathrm{eval}}}a_{ij}w_{ij}}
{\sum_{i\in\mathcal K_j^{\mathrm{eval}}}w_{ij}},
\qquad 0\le \mathrm{DEC}_j\le1,
\label{eq:diagnostic_evidence_coverage}
\end{equation}
when the denominator is positive, and $0$ otherwise. The implementation exports the same quantity as \texttt{Evidence\_Coverage} and \texttt{Confidence\_Score}. For interpretive reporting only, \texttt{Confidence\_Level} is High for $\mathrm{DEC}\ge0.80$, Moderate for $0.50\le\mathrm{DEC}<0.80$, and Low otherwise. These categories are evidence-completeness descriptors, not calibrated probabilities or validated clinical confidence cutoffs.

Crucially, $\mathrm{DEC}_j$ is not multiplied into $P_{\mathrm{evidence}}$, $r_j$, $c_j$, $R(D_j)$, or the hard-rule score. The exported $\texttt{Confidence\_Score}$ (numerically identical to DEC) is appended as the fourth coordinate of $S_0(D_j)$ for unsupervised partitioning; thus it can affect cluster geometry while remaining outside disease ranking and diagnostic evidence accumulation. The audit trace additionally lists unresolved Pathognomonic, Hallmark, Major, and Exclusionary EFNs so that equal numerical coverage arising from clinically different missing-evidence patterns remains distinguishable.
\end{bluereview}
\begin{bluereview}
The second patient--reference comparison is performed at the disease level and
is kept distinct from the feature-level Information Gate. For candidate
$D_j$, define the Hallmark/Major subset
\[
\mathcal H_j=\{i\in\mathcal O_j:g_{ij}\in\{0.85,0.70\}\},
\]
corresponding to Hallmark and Major diagnostic roles in the fixed ontology.
Collect patient and canonical magnitudes only over this subset,
\[
\mathbf p_j^{HM}=(p_{ij})_{i\in\mathcal H_j},
\qquad
\mathbf e_j^{HM}=(e_{ij})_{i\in\mathcal H_j},
\]
so that supportive/associated/nonspecific features can still contribute to
$E_{\rm net}(D_j)$ but do not determine aggregate disease-level similarity.
Pathognomonic features are handled separately by the hard-rule pathway. For
each retained feature,
\begin{equation}
e_{ij}=\|\mathbf r_{ij}\|_2=\sqrt{m_i}.
\label{eq:canonical_feature_magnitude}
\end{equation}
Here $p_{ij}=\|\mathbf q_{ij}\|_2$ is the activated patient-feature magnitude
from Eq.~\eqref{eq:pij_implementation}, whereas $e_{ij}$ is the corresponding
canonical feature magnitude. Disease-level directional agreement and relative
magnitude are then
\begin{equation}
c_j=
\frac{(\mathbf p_j^{HM})^{\top}\mathbf e_j^{HM}}
{\|\mathbf p_j^{HM}\|_2\,\|\mathbf e_j^{HM}\|_2},
\qquad
r_j=
\frac{\min(\|\mathbf p_j^{HM}\|_2,\|\mathbf e_j^{HM}\|_2)}
{\max(\|\mathbf p_j^{HM}\|_2,\|\mathbf e_j^{HM}\|_2)},
\label{eq:disease_similarity}
\end{equation}
with either quantity defined as $0$ when its denominator is zero. Thus the
first comparison, $IG_{ij}$, evaluates component-level correspondence within
one feature, whereas $(c_j,r_j)$ evaluates the aggregate patient--canonical
profile specifically across observed Hallmark/Major features of the same
candidate disease. The implementation deliberately does not collapse these quantities through a logistic similarity gate. Instead, it preserves the four-dimensional post-KG clustering representation
\begin{equation}
S_0(D_j)=
\begin{bmatrix}
P_{\mathrm{evidence}}(D_j)\\ r_j\\ c_j\\ \texttt{Confidence\_Score}(D_j)
\end{bmatrix},
\label{eq:s0_vector}
\end{equation}
which is the default full representation used by the reported four-coordinate experiments; the package also permits selected coordinate subsets. 

For candidate disease $D_j$, ranking compares the combined patient-derived
disease-level components with their corresponding canonical components.
The ranking score is defined as
\begin{equation}
R(D_j)
=
\frac{
P_{\mathrm{evidence},j}^{+}+r_j+c_j
}{
P_{\mathrm{evidence},j}^{\mathrm{can}}
+r_j^{\mathrm{can}}
+c_j^{\mathrm{can}}
},
\label{eq:ranking_score}
\end{equation}
where
\begin{equation}
P_{\mathrm{evidence},j}^{+}
=
\max\!\left(0,P_{\mathrm{evidence},j}\right).
\end{equation}
Here $P_{\mathrm{evidence},j}^{\mathrm{can}}$, $r_j^{\mathrm{can}}$, and $c_j^{\mathrm{can}}$ are the corresponding disease-level quantities obtained from the candidate's canonical self-reference under the same available-feature support. The denominator is therefore derived from the candidate representation rather than imposed as an arbitrary constant. Thus, the three disease-level components contribute independently to the ranking score. A zero value in one component does not multiplicatively suppress informative values in the remaining components. The raw, potentially negative $P_{\mathrm{evidence}}$ is retained unchanged in $S_0$; lower truncation is used only for its contribution to candidate ranking. The resulting $R(D_j)$ is a relative diagnostic ranking score, not a calibrated posterior disease probability.
\end{bluereview}

\subsubsection{Hard clinical rules as implemented}

\begin{bluereview}
A candidate $D_j$ triggers pathognomonic evidence when an observed/evaluable
feature satisfies $\exists i\in\mathcal O_j:g_{ij}=1,\ IG_{ij}\ge\tau_p$ and
triggers exclusionary evidence when
$\exists i\in\mathcal O_j:g_{ij}=-.5,\ IG_{ij}\ge\tau_e$, with
$\tau_p=\tau_e=.8$. Because $IG_{ij}$ is candidate-conditioned, a trigger
requires patient--reference agreement with the canonical state of that
candidate disease rather than mere presence of the raw patient variable.
\end{bluereview}

The saved implementation applies candidate-level pathognomonic/exclusionary overrides as formalized in Eq.~\eqref{eq:final_score}; candidates are scored before the maximum is selected, so a trigger is not global early termination, and its rule trace is reported.

\begin{bluereview}
The engine implements both pathognomonic and exclusionary candidate-level overrides. In the supplied knowledge files used by the six executed notebooks, however, no evaluated Dengue, Malaria, or Influenza EFN is encoded with \texttt{diagnostic\_role: Exclusionary}; consequently, the exclusionary override is available in code but is not empirically triggered in these six runs. In Dataset A, the raw column \texttt{MAL} denotes \emph{malaise}; the Dengue manifestation YAML maps it to a Supportive feature rather than an exclusionary malaria indicator. Role metadata must also be interpreted jointly with the candidate-specific canonical state: NS1 is Pathognomonic in the laboratory knowledge models, but its expected state differs by disease. Feature name/role alone is therefore not disease-specific evidence; the complete canonical representation and candidate-level quantities are.
\end{bluereview}

\subsubsection{Candidate ranking, decisive rules, and evidence-completeness audit}
Before candidate-level hard-rule override, diseases are ordered by the additive ranking score $R(D_j)$ from Eq.~\eqref{eq:ranking_score}. The post-rule score is
\begin{equation}
\Sscore(D_j)=
\begin{cases}
1,&\text{pathognomonic trigger},\\
0,&\text{otherwise, exclusionary trigger},\\
R(D_j),&\text{otherwise}.
\end{cases}
\label{eq:final_score}
\end{equation}
The leading represented hypothesis is $D^*=\arg\max_j\Sscore(D_j)$. Optional normalized support $P(D_j)=\Sscore(D_j)/\sum_r\Sscore(D_r)$ is a relative ranking weight and is not interpreted as a calibrated disease probability. Candidate-level pathognomonic and exclusionary rules remain explicit overrides rather than learned parameters.

Evidence completeness is audited separately from candidate scoring. For candidate $D_j$, Diagnostic Evidence Coverage ($\mathrm{DEC}_j$) is the weighted fraction of cohort-evaluable canonical diagnostic evidence observed for the patient. The exported Confidence Score (CS) is the corresponding evidence-completeness quantity used for reporting and as the fourth coordinate of the post-reasoning representation $S_0(D_j)=[P_{\mathrm{evidence}}(D_j),r_j,c_j,CS_j]^\top$. CS does not enter $P_{\mathrm{evidence}}$, $r_j$, $c_j$, or the additive ranking score $R(D_j)$, but it can affect the diagnostic assignments produced by K-means because it is part of $S_0$. Unresolved Pathognomonic, Hallmark, Major, and Exclusionary EFNs are exported explicitly. Missing evidence is therefore distinguished from observed absence without being converted into contradictory evidence.

\begin{algorithm}[t]
\footnotesize
\caption{CKG Reasoner: Inference and Diagnostic Assignment}
\label{alg:hybrid}

\begin{algorithmic}[1]
\Require Frozen KG, cohort $\mathcal X$,
candidate diseases $\mathcal D$, target $D_t$, $K=2$
\Ensure Inference traces $\mathcal T$,
rankings $\mathcal R$, diagnostic assignments $\hat{\mathbf y}$

\Statex \textbf{Phase I: Knowledge-grounded inference}

\For{$x\in\mathcal X$}
    \For{$D_j\in\mathcal D$}

        \For{$F_i\in\mathcal O_j(x)$}
            \State $\mathbf q_{ij}
            \gets \mathcal M_j(x,F_i)$
            \State $IG_{ij}
            \gets
            \sqrt{(\cos^2\theta_{ij}+\rho_{ij}^2)/2}$
            \State $(\texttt{pcon}_{ij},
            \texttt{ncon}_{ij})
            \gets \operatorname{Evidence}(IG_{ij},EFN_{ij})$
        \EndFor

        \State $P_{\mathrm{evidence},j}
        \gets E_{\mathrm{net},j}/E_{\mathrm{can},j}$
        \State $(r_j,c_j)
        \gets \operatorname{Similarity}
        (\mathbf p_j^{HM},\mathbf e_j^{HM})$
        \State $R_j
        \gets \operatorname{Rank}
        (P_{\mathrm{evidence},j},r_j,c_j)$
        \State $\mathcal S_j
        \gets \operatorname{Rules}(R_j,IG_j,EFN_j)$

        \State $CS_j
        \gets \mathrm{DEC}_j(x)$

        \State $S_0(x,D_j)
        \gets [P_{\mathrm{evidence},j},
        r_j,c_j,CS_j]^\top$

        \State $\mathcal T_{x,j}
        \gets \operatorname{Trace}(x,D_j)$

    \EndFor

    \State $\mathcal R_x
    \gets \operatorname{argsort}_{D_j}
    (\mathcal S_j)$

\EndFor

\Statex \textbf{Phase II: Unsupervised assignment}

\State $Z
\gets \{S_0(x,D_t):x\in\mathcal X\}$
\State $\widetilde Z
\gets \operatorname{Standardize}(Z)$
\State $\mathbf z
\gets \operatorname{KMeans}(\widetilde Z,K)$
\State $\hat{\mathbf y}
\gets \operatorname{OrderAndAssign}(\mathbf z,Z)$

\Statex \textbf{Phase III: Retrospective evaluation}

\State $\mathcal E
\gets \operatorname{Evaluate}
(\hat{\mathbf y},\mathbf y)$

\State \Return $\mathcal T,\mathcal R,
\hat{\mathbf y},\mathcal E$

\end{algorithmic}
\end{algorithm}

The clinical attributes used by the inference procedure are defined together in Section~\ref{sec:unified_attributes}; the operational scoring section above specifies how the patient and reference quantities enter those attributes. No separate interpretation table is repeated here.

\subsubsection{Patient-specific graph extraction and inference-faithful explanation}
For patient evidence $O=\{o_1,\ldots,o_n\}$, the system retrieves disease-specific EFNs and constructs $\mathcal G_P\subseteq\mathcal G_R$. The same observation may enter several candidate graphs with different canonical representations and evidence values; the dengue experiments use the dengue-oriented configuration only. Explainability is intrinsic because each candidate trace exposes the same patient/reference agreement, separated evidence attributes, decisive-rule states, and score that directly produce support, with provenance to the activated EFNs rather than a post-hoc surrogate. The EFNs alone are therefore not treated as complete explanations; an explanation is the patient-activated inference trace assembled from those EFNs and the similarity, evidence, rule, uncertainty, decision, and provenance quantities generated during inference.

\subsubsection{Formal explanation and reasoning-to-language extension}
\label{sec:fol_explanation}
The executed six-cohort experiments evaluate the knowledge-grounded
scoring, evidence audit, and partition-derived assignments; they do not
execute or validate FOL-based inference or LLM-generated clinical reports.
The package includes explanation-object construction, FOL-compatible
serialization methods, deterministic report generation, and consistency
checks as software interfaces. These facilities establish an avenue for
subsequent clinically grounded explanation research, not evidence of
experimentally demonstrated FOL/LLM explanation performance.

A proposed extension represents the patient-specific
inference trace for patient $x$ and candidate $D_j$ as
\begin{equation}
\mathcal Z_{x,j}=(\mathcal F_{x,j},\mathcal C_{x,j},\mathcal R_{x,j},
\mathcal U_{x,j},\mathcal D_{x,j},\Pi_{x,j}),
\label{eq:explanation_object}
\end{equation}
where the components organize evidence facts, numerical
contributions, rule states, evidence completeness, candidate decisions,
and provenance. FOL-compatible predicates can express observed evidence,
importance, specificity, similarity, positive and negative contributions,
rule triggers, evidence coverage, and candidate rank. Numerical operations
remain external arithmetic; this is not a claim that diagnostic scoring
or clustering is performed by pure first-order logic. The proposed
explanation object should distinguish candidate ranking from the
partition-derived diagnostic assignment.

\subsubsection{Prospective constrained clinical-report generation}
\label{sec:report_generation}
A future LLM-based verbalizer could receive only a
structured, provenance-linked explanation object and a fixed report
schema, with generation restricted to claims licensed by that object.
A deterministic verifier could check disease identity, numerical values,
support/contradiction signs, rule states, evidence completeness, and the
separate partition assignment; unsupported statements would require
rejection or deterministic fallback. The package's existing deterministic
report and verification utilities are preparatory components. No LLM
verbalizer, end-to-end claim-grounding evaluation, clinician-rated
explanation study, or clinical report-quality result is presented here.

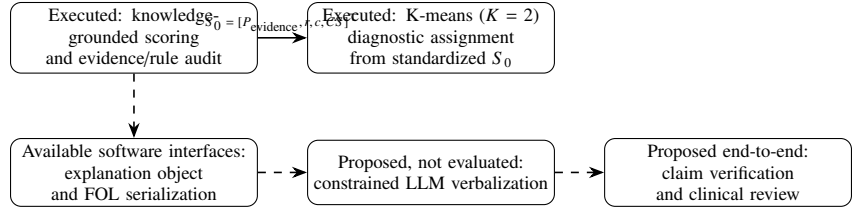
\begin{figure*}[t]
\centering
\scriptsize
\begin{tikzpicture}[
 >=Stealth, node distance=0.55cm,
 box/.style={draw,rounded corners,align=center,text width=3.1cm,minimum height=0.85cm},
 flow/.style={->,line width=0.6pt},
 opt/.style={->,dashed,line width=0.6pt}]
\node[box] (infer) {Executed: knowledge-grounded scoring\\and evidence/rule audit};
\node[box,right=0.65cm of infer] (cluster) {Executed: K-means ($K=2$)\\diagnostic assignment\\from standardized $S_0$};
\node[box,below=0.85cm of infer] (trace) {Available software interfaces:\\explanation object and FOL serialization};
\node[box,right=0.65cm of trace] (gen) {Proposed, not evaluated:\\constrained LLM verbalization};
\node[box,right=0.65cm of gen] (verify) {Proposed end-to-end:\\claim verification and clinical review};
\draw[flow] (infer) -- node[above,font=\tiny,align=center] {$S_0=[P_{\mathrm{evidence}},r,c,CS]^\top$} (cluster);
\draw[opt] (infer) -- (trace);
\draw[opt] (trace) -- (gen);
\draw[opt] (gen) -- (verify);
\end{tikzpicture}
\caption{Executed inference and partition assignment versus
available explanation interfaces and proposed, unevaluated FOL/LLM-based
clinical-report workflow. Dashed arrows denote extension paths rather
than experimental results.}
\label{fig:explanation_layer}
\end{figure*}

\subsubsection{End-to-end diagnostic workflow}
\begin{figure*}[t]
\centering
\scriptsize
\begin{tikzpicture}[
    >=Stealth,
    node distance=0.55cm,
    box/.style={draw,rounded corners,align=center,text width=2.65cm,minimum height=0.85cm},
    flow/.style={->,line width=0.6pt}
]
\node[box] (obs) {Patient observations};
\node[box,right=0.45cm of obs] (efn) {Candidate-specific EFN activation};
\node[box,right=0.45cm of efn] (ev) {Separated positive/negative evidence};
\node[box,right=0.45cm of ev] (rank) {Candidate score $R(D_j)$ + hard-rule override};
\node[box,right=0.45cm of rank] (audit) {Evidence audit: $\mathrm{DEC}_j$, CS, unresolved EFNs};
\draw[flow] (obs) -- (efn);
\draw[flow] (efn) -- (ev);
\draw[flow] (ev) -- (rank);
\draw[flow] (rank) -- (audit);
\node[box,below=0.85cm of rank] (part) {K-means diagnostic assignment ($K=2$)\\from standardized derived\\$S_0=[P_{\mathrm{evidence}},r,c,CS]^\top$};
\draw[flow] (audit.south) |- (part.east);
\end{tikzpicture}
\caption{Cross-sectional reasoning and evaluation workflow. Patient observations activate candidate-specific EFNs, graded evidence is accumulated, and candidate ranking remains distinct from the evidence-completeness audit. The audit reports $\mathrm{DEC}_j$, CS, and unresolved critical evidence. Post-reasoning K-means partitioning produces the reported cohort-level diagnostic tags. It operates on the standardized representation $S_0=[P_{\mathrm{evidence}},r,c,CS]$ and does not modify candidate evidence accumulation or ranking.}
\label{fig:workflow}
\end{figure*}
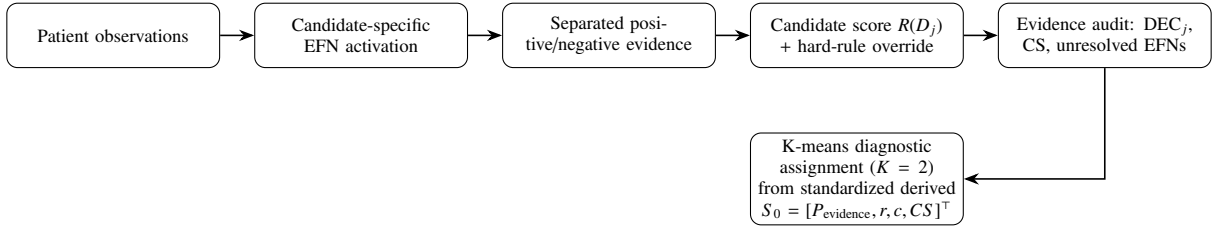

Figure~\ref{fig:workflow} summarizes
\[
\begin{aligned}
\text{KG}
&\rightarrow \text{EFNs}
\rightarrow \text{separated evidence}
\rightarrow \text{candidate activation} \\
&\rightarrow \text{graded aggregation}
\rightarrow \text{hard rules}
\rightarrow \text{candidate ranking} \\
&\rightarrow \text{evidence-completeness audit} \\
&\rightarrow \text{standardized } S_0=[P_{\mathrm{evidence}},r,c,CS]^\top \\
&\rightarrow \text{K-means diagnostic assignment }(K=2).
\end{aligned}
\]
The executed pathway produces the knowledge-grounded trace and K-means partition assignments. Conversion to a formal explanation object and any verified LLM report are extension paths, not operations evaluated in these experiments. Cohort A remains dengue versus non-dengue and does not relabel negative records as other diseases.

\subsection{Cross-Sectional One-Layer Formulation}
The dengue data contain one cross-sectional record per patient, so no temporal progression is inferred. The patient graph is $\mathcal G_P=(V_P,E_P)$ with $V_P\subseteq L_1$; feature evidence is aggregated, canonically normalized, and then passed through the disease-similarity and hard-rule mechanisms above. Because diagnostic-time attributes are absent, the patient score is a diagnostic evidence score (\texttt{Best\_Score}), not a time-normalized DEI.

\subsection{Training-Free Multi-Dataset Evaluation}
\label{sec:evaluation}

\subsubsection{Evaluation principle}
\begin{bluereview}
The evaluation uses six executed notebooks and the CKG Reasoner package, which was developed and frozen independently of all six datasets. Dataset-specific schema mapping and label-free cohort-level fitting of the integrated diagnostic-assignment stage are evaluation operations, not package construction or knowledge-model fitting. Outcome labels are not supplied to \texttt{hmap}, \texttt{graphfit}, candidate scoring, the ranking function, or K-means fitting. Each notebook performs deterministic schema mapping and exhaustive reasoning over the three bundled disease graphs. For target-disease evaluation, each patient is represented by
\begin{equation}
S_0(D_j)=\left[P_{\mathrm{evidence}}(D_j),r_j,c_j,\texttt{Confidence\_Score}(D_j)\right]^\top,
\end{equation}
whose four coordinates are standardized before K-means. The scalar ranking score $R(D_j)$ is not used as a clustering coordinate. Ground-truth labels enter only after cluster assignment for retrospective metric calculation. Ground-truth outcomes are therefore external retrospective criteria used to characterize cluster composition; they are not used to construct $S_0$, fit K-means, estimate centroids, or tune the reasoning parameters.

K-means is an integral component of the proposed diagnostic-assignment methodology, not a post-hoc analysis. It operates exclusively on the four knowledge-grounded evidence coordinates derived by the fixed reasoner; neither raw clinical predictors nor target/outcome labels are used as clustering inputs. K-means is the integrated diagnostic-assignment stage following the internal KG evidence and hard-rule pathway. As a subsequent stage, it does not modify EFNs, Information Gates, evidence contributions, hard-rule states, or candidate ranking. Nevertheless, because its assignments are used in the headline retrospective classification metrics, the uniform $K=2$ configuration is the diagnostic-assignment protocol evaluated here, not a clinically validated decision rule. It partitions the post-KG evidence representation rather than serving as a separate exploratory analysis or a clinically validated decision threshold. It examines whether the structured representation contains recoverable cohort-level diagnostic information without outcome-label fitting; the candidate-conditioned evidence pathway, hard clinical roles, ranking score, and evidence-completeness audit remain mathematically distinct from this integrated diagnostic-assignment stage.
\end{bluereview}

\begin{bluereview}\noindent The primary evaluation examines label-free separation of the fixed, knowledge-grounded four-coordinate representation under a uniform $K=2$ protocol. Neither raw dataset predictors nor target/outcome labels enter K-means; outcomes are used only afterward to quantify agreement with the resulting assignments.\end{bluereview}

\subsubsection{Evaluation datasets}
\begin{bluereview}
The evaluation comprises six cohorts spanning three target diseases (Table~\ref{tab:cohorts}). The first four reported evaluation cohorts use the package schemas \texttt{C1\_Bangladesh\_Dengue\_1000}, \texttt{C2\_D4\_Dengue\_Hematology\_1523}, \texttt{C3\_M1\_Malaria\_2190}, and \texttt{C4\_I1\_Thailand\_Influenza\_4569}. The two subsequently reported dengue cohorts likewise use the same frozen package and existing Dengue knowledge representation through deterministic schema mapping. Ground-truth positive/negative counts are 533/467, 1042/481, 1068/1122, 1493/3076, 644/345, and 697/321 for A--F, respectively.

\begin{table*}[t]
\centering
\caption{Datasets used in the executed CKG Reasoner evaluation.}
\label{tab:cohorts}
\scriptsize
\begin{tabularx}{\textwidth}{p{0.08\textwidth}p{0.10\textwidth}p{0.27\textwidth}p{0.17\textwidth}X}
\toprule
\textbf{Dataset} & \textbf{Target} & \textbf{Package schema} & \textbf{N / outcome} & \textbf{Evidence profile}\\
\midrule
A & Dengue & \texttt{C1\_Bangladesh\_Dengue\_1000} & 1000; 533 positive / 467 negative & Mixed serological, clinical, and routine laboratory evidence\\
B & Dengue & \texttt{C2\_D4\_Dengue\_Hematology\_1523} & 1523; 1042 positive / 481 negative & Hematology-dominant evidence\\
C & Malaria & \texttt{C3\_M1\_Malaria\_2190} & 2190; 1068 positive / 1122 negative & Malaria-specific clinical/laboratory evidence\\
D & Influenza & \texttt{C4\_I1\_Thailand\_Influenza\_4569} & 4569; 1493 positive / 3076 negative & Symptoms plus molecular PCR and rapid-antigen evidence\\
E & Dengue & D7 additional evaluation & 989; 644 positive / 345 negative & Hematology-focused; four headings mapped to existing Dengue EFNs\\
F & Dengue & D3 additional evaluation & 1018; 697 positive / 321 negative & Clinical symptoms plus platelet and WBC evidence; eight headings mapped to existing Dengue EFNs\\
\bottomrule
\end{tabularx}
\end{table*}
\end{bluereview}

\subsubsection{Disease-specific feature mapping}
Each dataset variable is mapped to a disease-specific EFN before outcome evaluation. Mapping is deterministic and semantic rather than outcome-fitted. The mapping layer preserves stage-specific EFN identity, normalizes supported units/categorical states, and distinguishes missing values from observed negative findings. In the influenza mapper, molecular assay subtype strings are normalized to \texttt{Influenza\_A\_or\_B}; rapid-antigen variants such as \texttt{Positive FluA}, \texttt{Positive FLU A}, \texttt{Positive FluB}, and combined A+B forms are normalized to \texttt{Positive}, whereas invalid or pending results are treated as unavailable. The generalized-aches source variable maps to GBA rather than being duplicated into both GBA and MYA.

For the audited dengue cohort, the preserved mapping is shown in Table~\ref{tab:mapping}. For the two additional dengue evaluations using the same frozen package, D7 mapped hemoglobin, WBC count, platelet count, and platelet-distribution width to existing EFNs; differential count and RBC count were skipped as unmatched, while age, gender, identifier, and the outcome field were not used as evidence. D3 mapped platelet count, WBC count, fever, fever duration, headache, myalgia, rash, and vomiting; identifier, gender, age, and the outcome field were not used as evidence. No outcome labels were supplied to the mapping or reasoning steps.
\begin{table}[t]
\centering
\caption{Audited mapping of variables in Dengue Cohort A to ontology features.}
\label{tab:mapping}
\scriptsize
\setlength{\tabcolsep}{3pt}
\begin{tabular}{llll}
\toprule
\textbf{Original variable} & \textbf{Ontology feature} & \textbf{Original variable} & \textbf{Ontology feature}\\
\midrule
Gender & Gender & Age & Age\\
NS1 & NS1 & IgG & IGG\\
IgM & IGM & Fever Duration & FEV\_dura\_dy\\
Body Temperature & FEV\_temperature\_c & Platelet Count & MTP\_platelet\_count\\
WBC Count & LEU\_wbc & Joint Pain & ART\_severity\\
Headache & HDH & Retro-Orbital Pain & ROP\\
Myalgia & MYA & Rash & RSH\\
\bottomrule
\end{tabular}
\end{table}

\subsubsection{Primary endpoints and partition protocol}
\begin{bluereview}
The main F1 is conventional positive-class F1 (\texttt{f1}, identical to \texttt{f1\_positive}); additional outputs include accuracy, balanced accuracy, macro/weighted F1, MCC, Cohen's $\kappa$, confusion matrices, ROC--AUC, average precision where available, and coverage. All six reported evaluations use $K=2$ on the standardized four-coordinate post-reasoning representation. K-means fitting does not use reference outcomes. The headline metrics are drawn from the recorded uniform $K=2$ evaluations. Because two clusters are assigned negative and positive tags, no intermediate partition is generated and partition-decision coverage is 1.000 in all six cohorts. The uniform retrospective protocol was not prospectively preregistered.
\end{bluereview}

\subsubsection{Evidence-completeness reporting}
For every prediction, the output includes patient-level Evidence Coverage, CS, confidence level/reason, and disease-specific unresolved critical-evidence fields, together with feature-level evidence and explanation traces. These quantities audit the completeness of the evidence available to the reasoner. They do not alter feature-level evidence accumulation or the candidate-ranking score. CS is, however, the fourth standardized coordinate of $S_0$ and can therefore affect K-means diagnostic assignments and their evaluation metrics.

\subsection{Comparators and Controlled Experiments}

\subsubsection{Comparator scope}

In addition to the knowledge-based comparator feasibility assessment
in Table~\ref{tab:reproducibility_audit6}, a supervised logistic-regression baseline table is supplied for all six datasets in
Table~\ref{tab:logistic_regression_all_cohorts}. This provides a contextual
predictive reference rather than a matched evaluation of the training-free
reasoning architecture. The logistic-regression results are based on the
reported held-out test subsets, whereas the CKG results characterize
K-means diagnostic assignments from post-reasoning representations. The comparison
therefore does not establish predictive superiority of either approach.

To examine the feasibility of comparative evaluation,
we further assessed the implementation availability
and adaptation requirements of representative diagnostic
reasoning systems identified in the related-work analysis
(Table~\ref{tab:relatedcomparison}).
The assessment considered the availability of executable
implementations, supporting model artifacts, input-data
requirements, and the compatibility of each architecture
with the structured cross-sectional datasets used in
the present study. The findings are summarized in
Table~\ref{tab:reproducibility_audit6}.

This assessment distinguishes the availability of a
published implementation from its direct applicability
to the target cohorts. Several existing approaches
operate on clinical narratives, specialized ontologies,
learned graph representations, or disease-specific
knowledge structures that differ substantially from
the available clinical and laboratory variables.
Consequently, reproducing their published results
and adapting them to the present datasets constitute
distinct experimental tasks. The assessment does not
imply that these architectures are intrinsically
irreproducible or that their adaptation is impossible.

\begin{table*}[t]
\centering
\caption{Reproducibility and empirical execution audit of representative diagnostic reasoning systems surveyed in Table~\ref{tab:relatedcomparison} when considered for baseline benchmarking on our target cross-sectional cohorts.}
\label{tab:reproducibility_audit6}
\scriptsize
\setlength{\tabcolsep}{4pt}
\renewcommand{\arraystretch}{1.18}
\begin{tabularx}{\textwidth}{p{0.09\textwidth}p{0.24\textwidth}p{0.14\textwidth}X}
\toprule
\textbf{Study} & \textbf{Method / Architecture Description} & \textbf{Artifact / Access Status} & \textbf{Technical Reason Precluding Execution on Target Cohorts} \\
\midrule
\cite{DEDUP_01384} & Hybrid rule-based/probabilistic expert system combining expert outputs via neural network. & Paywalled (IEEE) \newline No public code & Historical 1992 proceeding pre-dating digital code repositories; no source code, inference software, or calibrated parameters available. \\
\addlinespace[2pt]
\cite{DEDUP_00138} & Ontology-grounded fuzzy decision support system using semantic similarity for diabetes. & Open Access (IEEE Access) \newline No public code & Code unreleased; hospital training dataset is private; transfer to acute febrile cohorts impossible without calibrated fuzzy membership functions. \\
\addlinespace[2pt]
\cite{SB_DEDUP_00169} & Siamese Bayesian networks incorporating symptom absence as negative evidence over learned BNs. & Paywalled (ACM) \newline Proprietary commercial IP & Developed as proprietary commercial IP for the \textit{mFine} telemedicine platform; network topologies, learned weights, and code are strictly unreleased. \\
\addlinespace[2pt]
\cite{DEDUP_01201} & DKDR: Knowledge graph reasoning with deep reinforcement learning for interactive diagnosis. & Paywalled (IEEE) \newline No public code & Closed conference proceeding; no public code repository, pre-trained policy checkpoints, or KG simulation environment released. \\
\addlinespace[2pt]
\cite{SB_DEDUP_00479} & Multi-hop KG path reasoning with dynamic TF-IDF and Naive Bayes weights for TCM diagnosis. & Open Access (CMC) \newline No public code & Open-access publication, but no source code, reasoning scripts, or path weight tables were publicly deposited. \\
\addlinespace[2pt]
\cite{DEDUP_00010} & Semantic KG reasoning with likelihood-ratio-weighted edges and conditional patient constraints. & Paywalled (Springer) \newline No public code & Published as a conceptual book chapter in conference proceedings; no software implementation, graph exports, or code packages released. \\
\addlinespace[2pt]
\cite{DEDUP_03285} & ICHD-3 headache KG engine using weighted criteria matching and exclusion penalties. & Paywalled (Wiley) \newline Meeting abstract only & Conference meeting abstract only; no executable differential engine, ontology rules, or software packages were distributed. \\
\addlinespace[2pt]
\cite{DEDUP_00290} & D$^2$KGMed: Dynamic diagnostic knowledge graphs generated and refined via fine-tuned LLM. & Paywalled (IEEE) \newline  Closed conference proceeding; & No source code, dynamic graph generation routines, or fine-tuning weights released. \\
\addlinespace[2pt]
\cite{SC_03} & DR.KNOWS: Stack-GIN graph neural network over UMLS SNOMED-CT with multi-head attention path rankers. & Open Access (JMIR AI) \newline \textbf{Public repository exists} \newline (\texttt{serenayj/DRKnows}) & \textbf{Cannot be run out-of-the-box:} (1) Expects unstructured text notes and discrete UMLS CUIs rather than continuous laboratory measurements; (2) Pre-trained neural checkpoints (\texttt{gmodel.pth}, \texttt{encoder.pth}) were omitted and point to private author cluster paths; (3) Requires supervised training on credentialed MIMIC-III ICU data. \\
\addlinespace[2pt]
\cite{DEDUP_01211} & ReCLLaMA: Neuro-symbolic LLM agent linking procedure codes to proteins with NARS logic. & Conference proc. (IEEE) \newline \textbf{Public repository exists} \newline (\texttt{shilab/RECLLAMA}) & \textbf{Cannot be run on target cohorts:} (1) Internal reasoning graph (\texttt{diseases\_reasons.pickle}) has only 258 conditions, completely omitting Dengue (\texttt{061}), Malaria (\texttt{084}), and Influenza (\texttt{487}); (2) Alignment model requires surgical procedure codes to link proteins; (3) Missing local fine-tuned BioBERT weights. \\
\midrule
\textbf{Proposed} & Disease-specific EFNs + patient evidence graph + symbolic--probabilistic accumulation + hard rules. & \textbf{Source package and six executed notebooks supplied} & \textbf{Executed retrospectively on six cohorts; clinical validity remains unestablished;} deterministic, training-free, requires no unreleased checkpoints or outcome-fitting. \\
\bottomrule
\end{tabularx}
\end{table*}

\begin{table*}[t]
\centering
\caption{Cross-cohort supervised logistic regression baseline performance
across the six evaluation cohorts. Thailand Influenza and Dengue Cohort A
incorporate point-of-care rapid testing alongside clinical and laboratory
presentation. Bold highlights values as formatted in the supplied baseline table; cross-cohort values are not directly comparable because cohorts differ.}
\label{tab:logistic_regression_all_cohorts}
\scriptsize
\begin{tabular}{llrrrrrrrr}
\toprule
Cohort & Dataset Name & Accuracy & F1 Score &
Sensitivity & Specificity & Precision &
ROC-AUC & PR-AUC & Test Size (Pos / Neg) \\
\midrule
A
& Bangladesh mixed evidence (Dengue) 
& \textbf{100.00\%}
& \textbf{100.00\%}
& \textbf{100.00\%}
& \textbf{100.00\%}
& \textbf{100.00\%}
& \textbf{1.0000}
& \textbf{1.0000}
& 200 (107 / 93) \\

 B
& Hematology (Dengue)
& 60.00\%
& 68.23\%
& 62.68\%
& 54.17\%
& 74.86\%
& 0.6513
& 0.7793
& 305 (209 / 96) \\

 C
& Clinical Data (Bangladesh) (Malaria)
& 70.09\%
& 69.89\%
& 71.03\%
& 69.20\%
& 68.78\%
& 0.7539
& 0.7022
& 438 (214 / 224) \\

 D
& Thailand ILI cohort (Influenza)
& 88.84\%
& 82.59\%
& 80.94\%
& 92.68\%
& 84.32\%
& 0.9268
& 0.8784
& 914 (299 / 615) \\

 E
& D7 additional evaluation (Dengue)
& 92.42\%
& 94.30\%
& 96.12\%
& 85.51\%
& 92.54\%
& 0.8968
& 0.8918
& 198 (129 / 69) \\

 F
& D3 additional evaluation (Dengue)
& 99.02\%
& 99.29\%
& 99.29\%
& 98.44\%
& 99.29\%
& 0.9941
& 0.9971
& 204 (140 / 64) \\
\bottomrule
\end{tabular}\par\vspace{2pt}\footnotesize Note: Dataset E corresponds to D7 and Dataset F to D3. Baseline values are reported as verified by the authors; baseline and CKG evaluation protocols differ.
\end{table*}

\subsubsection{Ablation and robustness protocol}
The available secondary component ablations for B, C, D, and F remove $P_{\mathrm{evidence}}$, $r$, $c$, or CS from the post-KG partition representation, evaluate $P_{\mathrm{evidence}}$ alone and $(r,c)$ alone, remove the contradiction channel, and set clinical-importance weights to unity. Influenza additionally removes PCR alone and PCR together with rapid-antigen evidence. Evidence-withholding stress tests remove fixed fractions (10\%, 25\%, and 50\%) of observed evidence without parameter refitting. Evidence-gate sensitivity varies one parameter family at a time over $\alpha\in\{0.025,0.05,0.10\}$, $\beta\in\{5,10,20\}$, and $\gamma\in\{0.75,0.85,0.95\}$. Initialization stability is assessed across 20 K-means random seeds. Adjusted Rand index (ARI) measures partition agreement with the corresponding baseline. Bootstrap intervals and secondary stress-test results are reported only for cohorts with compatible $K=2$ configurations. These analyses do not refit ontology weights or activation parameters to outcome labels.
No quantitative evidence-acquisition or blinded explanation-quality experiment is reported because the supplied executed analyses do not contain such completed evaluations. These remain prospective validation targets and are therefore not presented as results.

\section{Results}

For context, Table~\ref{tab:logistic_regression_all_cohorts}
reports the supplied supervised logistic-regression baseline results for
Datasets A--F, including the respective held-out test-set sizes.
These results are presented separately from the CKG retrospective
partition metrics in Table~\ref{tab:mainresults}. In particular, the
reported F1 values have different evaluation populations and decision
procedures and must not be interpreted as a direct head-to-head test.

\subsection{Post-KG partition performance}
\begin{bluereview}
Table~\ref{tab:mainresults} reports the corrected K-means evaluation after synchronizing partition ordering with the additive candidate-ranking strategy. K-means fitting itself remains outcome-label-independent and operates on standardized four-dimensional $S_0=[P_{\mathrm{evidence}},r,c,CS]$ coordinates; the scalar $R(D_j)$ is not a clustering coordinate. All six cohorts have full partition-decision coverage under the uniform $K=2$ evaluation.

\begin{table*}[t]
\centering
\caption{Uniform $K=2$ KG partition results. F1 is positive-class F1; all six cohorts have full partition-decision coverage. MCC is omitted where it was unavailable for the uniform protocol.}
\label{tab:mainresults}
\scriptsize
\begin{tabular}{llrrrrrrr}
\toprule
Dataset & Target & N & Accuracy & Bal. Acc. & F1$_+$ & MCC & Coverage & Silhouette\\
\midrule
A: Bangladesh mixed evidence & Dengue & 1000 & 0.996 & 0.996 & 0.996 & -- & 1.000 & 0.724\\
B: Hematology & Dengue & 1523 & 0.558 & 0.558 & 0.634 & 0.108 & 1.000 & 0.635\\
C: Clinical Data (Bangladesh) & Malaria & 2190 & 0.707 & 0.706 & 0.695 & 0.413 & 1.000 & 0.889\\
D: Thailand ILI cohort & Influenza & 4569 & 0.906 & 0.871 & 0.842 & 0.783 & 1.000 & 0.842\\
E: D7 additional evaluation & Dengue & 989 & 0.914 & 0.895 & 0.936 & -- & 1.000 & 0.870\\
F: D3 additional evaluation & Dengue & 1018 & 0.893 & 0.907 & 0.917 & 0.776 & 1.000 & 0.601\\
\bottomrule
\end{tabular}
\end{table*}

Under the uniform $K=2$ evaluation, Dataset A has all-record accuracy 0.996, balanced accuracy 0.996, positive-class F1 0.996, and full coverage. Dataset E has accuracy 0.914, balanced accuracy 0.895, positive-class F1 0.936, and full coverage. Dataset B yields accuracy 0.558, balanced accuracy 0.558, F1 0.634, and MCC 0.108. Dataset C yields accuracy 0.707, balanced accuracy 0.706, F1 0.695, and MCC 0.413. Dataset D yields accuracy 0.906, balanced accuracy 0.871, F1 0.842, and MCC 0.783. Dataset F yields accuracy 0.893, balanced accuracy 0.907, MCC 0.776, and F1 0.917. All six cohorts have full partition-decision coverage. MCC for A and E is not reported because corresponding values were unavailable for the uniform protocol.

The influenza result remains subject to confirmatory-evidence circularity: PCR exactly separates the reference outcome and is mapped as pathognomonic influenza evidence. The confirmatory-evidence ablation below therefore provides the more informative stress test of this evidence regime.

Taken together, these experiments provide a feasibility and robustness assessment of the outcome-label-independent reasoning architecture rather than a claim of predictive superiority. The reported CKG classification metrics evaluate K-means-derived diagnostic tags from the post-KG representation. They characterize diagnostic assignments derived exclusively from knowledge-grounded evidence coordinates; prospective clinical performance, calibration, and clinical utility are not evaluated.
\end{bluereview}

\subsection{Bootstrap uncertainty and seed stability}
\begin{bluereview}
One thousand nonparametric bootstrap resamples yielded 95\% intervals for all-record accuracy and positive-class F1 of 0.533--0.581 and 0.607--0.658 for Dataset B, 0.687--0.725 and 0.672--0.716 for Dataset C, and 0.897--0.914 and 0.827--0.856 for Dataset D, respectively. Dataset F yielded corresponding intervals of 0.874--0.912 and 0.901--0.933, with full coverage. Comparable bootstrap intervals under the uniform protocol are not reported for A and E.

In the available $K=2$ secondary experiments for B, C, D, and F, partitions were unchanged across 20 K-means random seeds (ARI=1.000). This establishes initialization stability for these executed tests, not independent validation.
\end{bluereview}

\subsection{Component ablation}
\begin{bluereview}
Table~\ref{tab:ablation} reports the available $K=2$ component ablations for B, C, D, and F relative to the full four-dimensional representation. Comparable component-ablation results under the uniform protocol are not available for A and E. In Dataset B, removing $P_{\mathrm{evidence}}$, $r$, or $c$ yielded F1 scores of 0.612, 0.660, and 0.607, respectively, while $P_{\mathrm{evidence}}$ alone yielded 0.694. Dataset C was largely invariant to single-coordinate removal, with F1 0.695 except when $c$ was removed (0.692). Dataset D was similarly stable under single-coordinate removal; $P_{\mathrm{evidence}}$ alone produced perfect retrospective separation, which should be interpreted in light of the confirmatory laboratory evidence. In Dataset F, removing $P_{\mathrm{evidence}}$ changed the partition more substantially (ARI=0.561), while removing $r$ or $c$ retained ARI 0.949 and 0.942; removing CS reproduced the full partition (ARI=1.000).

\begin{table*}[t]
\centering
\caption{Available $K=2$ component ablations for Datasets B, C, D, and F relative to the full four-dimensional post-reasoning representation. Values are positive-class F1. The No-CS column is the explicit three-dimensional $[P_{\mathrm{evidence}},r,c]$ ablation.}
\label{tab:ablation}
\scriptsize
\begin{tabular}{lrrrrrrr}
\toprule
Dataset & Full & No $P_{\mathrm{evidence}}$ & No $r$ & No $c$ & $P_{\mathrm{evidence}}$ only & Profile only & No CS\\
\midrule
B & 0.634 & 0.612 & 0.660 & 0.607 & 0.694 & 0.612 & 0.634\\
C & 0.695 & 0.695 & 0.695 & 0.692 & 0.695 & 0.695 & 0.695\\
D & 0.842 & 0.842 & 0.843 & 0.844 & 1.000 & 0.842 & 0.842\\
F & 0.917 & 0.925 & 0.928 & 0.930 & 0.931 & 0.925 & 0.917\\
\bottomrule
\end{tabular}
\end{table*}

Removing contradiction or setting clinical-importance weights to one left the Dataset-C and D partitions unchanged (ARI=1.000). Dataset B changed modestly (ARI=0.984 and 0.969, with F1 0.635 and 0.633); in Dataset F, the corresponding ARIs were 0.976 and 0.988, with F1 0.913 and 0.917. These results indicate that component utility depends on the evidence regime rather than showing a uniform benefit in every cohort.
\end{bluereview}

\subsection{Evidence-gate, partition, and missing-evidence sensitivity}
\begin{bluereview}
The available $K=2$ sensitivity analyses show limited dependence on evidence-gate settings. Varying $\alpha$, $\beta$, and $\gamma$ over the specified ranges produced unchanged partitions (ARI=1.000) in nearly all tested configurations for B, C, D, and F. The largest departure was Dataset B at $\beta=5$ (ARI=0.971; F1=0.628 versus 0.634 at baseline); at $\gamma=0.95$, its ARI was 0.997. Dataset F remained stable, with its largest departure at $\beta=5$ (ARI=0.984; F1 0.919 versus 0.917 at baseline).

The common $K=2$ protocol ensures that all six cohorts are evaluated using the same partition specification. It does not establish that two clusters are clinically optimal.

Controlled random evidence withholding reduced performance in several cohorts. At 50\% withholding, Dataset B had F1 0.691, balanced accuracy 0.503, and MCC 0.006, illustrating why F1 alone is insufficient under class imbalance. Dataset C fell to F1 0.447 and Dataset D to F1 0.498. Dataset F yielded F1 0.862, accuracy 0.811, balanced accuracy 0.782, MCC 0.563, and full coverage. Corresponding uniform-protocol withholding results are not reported for A and E. These tests support reporting balanced accuracy and MCC alongside F1.
\end{bluereview}

\subsection{Influenza confirmatory-evidence ablation}
\begin{bluereview}
Removing PCR alone did not change the Dataset-D partition (ARI=1.000; F1=0.842), indicating that other encoded evidence retained the same retrospective separation. Removing both PCR and rapid-antigen evidence, however, reduced accuracy from 0.906 to 0.539, balanced accuracy from 0.871 to 0.627, F1 from 0.842 to 0.555, and MCC from 0.783 to 0.262 (ARI versus baseline $=-0.038$). Accordingly, the primary influenza analysis is interpreted as \emph{post-test evidence integration}, because confirmatory laboratory evidence is available to the reasoner. The PCR-plus-antigen removal experiment is reported separately as a \emph{pre-test-like evidence-withholding stress test}; it is not equivalent to a prospectively designed pre-test prediction study. The high influenza performance therefore depends substantially on the confirmatory laboratory evidence regime as a whole and should not be interpreted as pre-test symptom-only prediction.
\end{bluereview}

\subsection{Clustering diagnostics and ranking separation}
\begin{bluereview}
The uniform $K=2$ partitioning operates on standardized $[P_{\mathrm{evidence}},r,c,CS]$ rather than the scalar disease-ranking score $R(D_j)$. Candidate ranking and cohort partitioning therefore remain mathematically and operationally distinct. The two partitions are ordered using their mean evidence-plus-profile score, without using reference outcomes for fitting.
\end{bluereview}

\section{Discussion}

The six-cohort evaluation demonstrates the behavior of a
frozen, outcome-label-independent clinical reasoning
architecture across heterogeneous evidence regimes.
The uniform $K=2$ protocol provides a consistent
experimental specification, with K-means applied exclusively to the four derived
evidence coordinates, without raw predictors or target labels. The substantial
cross-cohort variation in classification performance
requires interpretation in relation to evidence
availability, diagnostic specificity, and the
corresponding supervised baselines, rather than
being attributed exclusively to the reasoning
architecture.

\subsection{Training-free ontology-grounded reasoning}

CKG Reasoner integrates candidate-specific cognitive
mapping, explicit clinical evidence semantics,
symbolic--probabilistic evidence accumulation,
disease-profile similarity, decisive clinical rules,
and evidence-completeness auditing within a fixed
inference architecture.

Unlike outcome-fitted classification, the diagnostic
reasoner operates without estimating its parameters
from cohort outcome labels. Its explicit separation
of positive support, contradiction, clinical
importance, specificity, and patient--reference
correspondence makes individual evidence
contributions and their provenance inspectable.

Importantly, relative candidate ranking and
cohort-level diagnostic assignment are distinct
operations. Candidate ranking uses the
canonically normalized additive score, subject
to pathognomonic and exclusionary overrides.
The reported classification metrics instead
evaluate unsupervised partitions of the
four-dimensional post-reasoning representation
$S_0=[P_{\mathrm{evidence}},r,c,CS]^\top$.
CS contributes to clustering but remains
separate from candidate ranking.

Consequently, the present study investigates
the feasibility, coherence, auditability,
and evidence sensitivity of a knowledge-grounded
reasoning paradigm complementary to
outcome-fitted classifiers, rather than
establishing comparative predictive superiority.

\subsection{Cross-cohort variability and baseline context}

The positive-class F1 scores range from
0.634 to 0.996 across the six cohorts.
This variation must be interpreted against
their heterogeneous clinical evidence
and evaluation conditions.

The supplied logistic-regression results
in Table~\ref{tab:logistic_regression_all_cohorts}
provide relevant contextual evidence.
Their reported F1 scores for cohorts A--E
are 1.000, 0.682, 0.699, 0.826, and
0.943, respectively, compared with CKG
scores of 0.996, 0.634, 0.695, 0.842,
and 0.936.

Both sets of reported results exhibit
substantial cross-cohort variation.
This observation is consistent with the
importance of dataset-specific evidence
and task characteristics, although it
does not establish that these factors
fully explain the observed differences.

The evaluation protocols must remain
distinguished. Logistic regression is
outcome-fitted and evaluated on reported
held-out subsets, whereas CKG classification
uses cohort-fitted, label-free partitions.
Consequently, numerical differences
between their F1 scores cannot establish
matched predictive superiority or
equivalence. The verified Dataset-F logistic-regression baseline
corresponds to the D3 cohort.

Dataset A combines serological and
clinical dengue evidence, including
confirmatory information. Its high
retrospective performance is therefore
interpreted within this comparatively
informative evidence regime.

In contrast, Dataset B relies
predominantly on routine hematological
findings with limited disease specificity.
Its CKG F1 of 0.634 and the reported
logistic-regression F1 of 0.682
provide complementary descriptive
evidence of more limited discrimination
under their respective protocols.
Neither result establishes the cause
of the performance limitation.

The additional dengue cohorts extend
evaluation of the frozen reasoning
specification to different evidence
subsets. Dataset E achieves F1 0.936
using four mapped hematological
headings, while Dataset F achieves
F1 0.917 using eight mapped clinical
and laboratory headings. These results
demonstrate retrospective separability
under additional schemas without
outcome-based modification of the
knowledge representation. They do
not establish prospective or
external-site clinical validity.

Malaria yields F1 0.695, compared
with the reported logistic-regression
F1 of 0.699 under its separate
evaluation protocol. This provides
additional context for interpreting
the moderate discrimination observed
in that cohort.

Influenza achieves F1 0.842, but
its reference outcome is perfectly
aligned with PCR, which is also
encoded as pathognomonic evidence.
The primary result therefore reflects
post-test evidence integration rather
than independent pre-test prediction.
Its reported logistic-regression
F1 of 0.826 is likewise interpreted
within an evaluation regime containing
confirmatory laboratory information.

Overall, the results indicate that
the fixed reasoning specification
can produce substantially different
retrospective discrimination across
clinical evidence regimes. This
variation should not be conflated
with computational instability,
nor should the reported baselines
be treated as matched comparative
validation.

\subsection{Robustness and evidence dependence}

The available secondary experiments
help distinguish sensitivity to
algorithmic configuration from
dependence on clinical evidence.

For cohorts B, C, D, and F,
the reported $K=2$ partitions
were unchanged across 20 K-means
random seeds. Evidence-gate
sensitivity experiments likewise
showed predominantly stable
partitions over the tested parameter
ranges, with limited departures
in selected configurations.

These findings support initialization
stability and limited parameter
sensitivity within the executed
experiments. They do not establish
stability under population shift,
alternative knowledge representations,
or prospective deployment.
Comparable secondary results for
A and E are not reported where
unavailable.

Component ablations further
demonstrate that the contribution
of individual reasoning coordinates
depends on the evidence regime.
For Dataset B, using
$P_{\mathrm{evidence}}$ alone
produced higher retrospective F1
than the full representation.
Dataset F showed greater partition
sensitivity to removing
$P_{\mathrm{evidence}}$, whereas
removing CS reproduced its
full partition.

These observations caution against
assuming that every reasoning
component contributes equally
to discrimination in every cohort.
The architecture preserves
their distinct clinical meanings
rather than optimizing their
combination against evaluation
outcomes.

Evidence-withholding experiments
provide an additional qualification.
At 50\% withholding, the reported
F1 declined to 0.447 for malaria
and 0.498 for influenza.
Dataset B illustrates the
importance of complementary metrics:
its F1 was 0.691, but balanced
accuracy was 0.503 and MCC
was 0.006.

For influenza, removing PCR alone
left the reported partition unchanged,
whereas removing both PCR and
rapid-antigen evidence reduced
F1 from 0.842 to 0.555.
This establishes substantial
dependence on the combined
confirmatory-evidence regime
in the executed retrospective
experiment.

Taken together, the available
robustness analyses support
computational stability under
the tested configurations while
also demonstrating clinically
important evidence dependence.
These are complementary,
not contradictory, findings.

\subsection{Explainability, uncertainty,
and translational extensibility}

An important architectural contribution
is the preservation of clinically
distinct reasoning quantities rather
than their reduction to a single
opaque classification output.

The implemented inference trace
exposes patient--reference
correspondence, diagnostic roles,
positive and contradictory evidence,
decisive-rule states, candidate
scores, evidence completeness,
and provenance.

Missing-aware normalization and
Diagnostic Evidence Coverage
address complementary questions:
the former characterizes correspondence
using available evidence, whereas
the latter identifies how much
cohort-evaluable diagnostic capacity
was observed. CS is therefore
an evidence-completeness descriptor,
not a calibrated probability
of diagnostic correctness.

This separation also creates
architectural opportunities for
translational research. Explicit
unresolved critical evidence,
candidate-specific reasoning,
and modular decision stages
provide a foundation for
investigating uncertainty regions,
interactive evidence acquisition,
and patient-wise diagnostic
reassessment rather than
requiring unconditional binary
classification.

The inference-faithful trace
additionally provides a structured
foundation for clinically grounded
explanations and prospective
real-time decision-support
applications.

These opportunities arise from
the framework's design, but
must be distinguished from
experimentally established
capabilities. The present
six-cohort evaluation does not
validate interactive evidence
acquisition, explicit uncertainty
regions, real-time deployment,
FOL-based inference, or
LLM-generated clinical
explanations.

Future studies should separately
evaluate these capabilities,
including explanation faithfulness,
clinical grounding, selective
prediction, information-acquisition
utility, and prospective
clinical performance.

\section{Limitations}
\begin{bluereview}

\paragraph{Clinical knowledge and evidence representation.}
The framework depends on the accuracy and completeness
of encoded medical knowledge and the consistency of
clinical feature mappings. The effects of population-specific
reference ranges, measurement normalization, and
feature-activation boundaries require further investigation.
Although exclusionary reasoning is implemented, its
behavior has not been empirically evaluated in the
six studied cohorts.

\paragraph{Hierarchical and cooperative evidence reasoning.}
The current framework does not explicitly model hierarchical,
conditional, or cooperative interactions among diagnostic features.
Multiple Hallmark or Major findings may provide conflicting
evidence, while their combined diagnostic significance may differ
from their individual contributions. For example, discordant
platelet and leukocyte measurements may require joint clinical
interpretation rather than independent evidence accumulation.
Future extensions should incorporate clinically justified feature
dependencies, hierarchical relationships, composite evidence
rules, and mechanisms for resolving conflicting observations.

\paragraph{Knowledge updating and distribution shift.}
The fixed knowledge representations do not automatically adapt
to changes in clinical populations, laboratory practices, or
disease manifestations. External medical knowledge-graph
integration, evidence retrieval, and controlled ontology-updating
mechanisms should therefore be investigated to improve contextual
adaptation and support the identification of distribution shift
and potential confounding. Such updates would require provenance,
clinical verification, and version control to preserve the
auditability of the reasoning process.

\paragraph{Explicit uncertainty and information acquisition.}
The current cohort-level partitioning does not establish a
clinically validated uncertainty region. Future extensions
should explicitly distinguish insufficient evidence, conflicting
evidence, and uncertain diagnostic assignments. Rather than
forcing a decision, the reasoner should be able to abstain
and identify additional clinically relevant observations
required to resolve uncertainty. Evidence adequacy, contradiction,
and the clinical consequences of alternative decisions should
inform this mechanism rather than cluster geometry alone.

\paragraph{Clinical translation and explanation.}
Diagnostic Evidence Coverage measures the availability of
cohort-evaluable evidence rather than calibrated diagnostic
certainty. Although excluded from candidate ranking, its
Confidence Score is included in the clustering representation
and can influence K-means diagnostic assignments. Matched comparisons
with supervised classifiers, calibration, clinical utility,
and prospective selective-decision evaluation remain outstanding.
The present framework is cross-sectional; temporal reasoning
and FOL/LLM-based clinical explanations have not undergone
clinical validation.

\end{bluereview}

\section{Conclusion}
\begin{bluereview}
CKG Reasoner investigates a complementary, outcome-label-independent approach to cross-sectional clinical reasoning through explicit medical knowledge and candidate-specific cognitive mapping. Its architecture separates feature-level patient--reference correspondence, positive and contradictory evidence, missing-aware canonical normalization, Hallmark/Major disease-profile similarity, decisive clinical rules, candidate ranking, and evidence-completeness auditing. The reported diagnostic tags are produced separately by cohort-fitted K-means on the four-coordinate post-reasoning representation $[P_{\mathrm{evidence}},r,c,CS]^\top$; they are not direct classifications from the candidate-ranking score.

Using the same frozen package and disease-specific knowledge representations, the uniform $K=2$ retrospective evaluation covered six cohorts: four dengue datasets ($N=1000,1523,989,1018$), malaria ($N=2190$), and influenza ($N=4569$). Their respective positive-class F1 scores were 0.996, 0.634, 0.936, 0.917, 0.695, and 0.842, with full partition-decision coverage in all six. Differences across cohorts underscore the dependence of retrospective separation on the available evidence regime. Compatible secondary analyses characterize component dependence, evidence withholding, parameter sensitivity, and partition stability where available; corresponding uniform-protocol robustness estimates for A and E remain unreported. Influenza performance incorporates confirmatory testing and must not be interpreted as independent pre-test prediction.

The contribution is an auditable reasoning architecture that keeps knowledge-grounded scoring and evidence provenance distinct from cohort-level diagnostic assignment without fitting the scorer to cohort outcome labels. The reported results establish neither prospective clinical validity nor predictive superiority over outcome-fitted classifiers. Future work should prioritize independent and temporal validation, matched clinical-task comparisons, assessment of encoded knowledge and evidence interactions, calibrated or selective clinical decision procedures, and clinician evaluation of the proposed FOL/LLM explanation extension.
\end{bluereview}


\begin{thebibliography}{00}
\footnotesize
\setlength{\itemsep}{0pt}
\setlength{\parskip}{0pt}
\bibitem{DEDUP_01384} K. Henson-Mack, H.-C. Chen, D. C. Wester, Integrating probabilistic and rule-based systems for clinical differential diagnosis, Proceedings IEEE Southeastcon '92 (1992) 699--702. \url{https://doi.org/10.1109/SECON.1992.202287}.
\bibitem{DEDUP_01400} K. C. C. Chan, J. Y. Ching, A. K. C. Wong, A probabilistic inductive learning approach to the acquisition of knowledge in medical expert systems, Proceedings Fifth Annual IEEE Symposium on Computer-Based Medical Systems (1992) 572--581. \url{https://doi.org/10.1109/CBMS.1992.245017}.
\bibitem{DEDUP_01222} P. Agarwal, R. Verma, A. Mallik, Ontology based disease diagnosis system with probabilistic inference, 2016 1st India International Conference on Information Processing (2016) 1--5. \url{https://doi.org/10.1109/IICIP.2016.7975383}.
\bibitem{DEDUP_00123} D. K. Choubey, S. Paul, V. K. Dhandhenia, Rule based diagnosis system for diabetes, Biomedical Research (2017) 5196--5209.
\bibitem{DEDUP_00162} N. Shoaip, S. El-Sappagh, S. Barakat, M. Elmogy, Ontology enhanced fuzzy clinical decision support system, U-Healthcare Monitoring Systems: Volume 1: Design and Applications (2018) 147--177. \url{https://doi.org/10.1016/B978-0-12-815370-3.00007-4}.
\bibitem{DEDUP_00138} S. El-Sappagh, J. M. Alonso, F. Ali, A. Ali, J.-H. Jang, K.-S. Kwak, An ontology-based interpretable fuzzy decision support system for diabetes diagnosis, IEEE Access (2018) 37371--37394. \url{https://doi.org/10.1109/ACCESS.2018.2852004}.
\bibitem{SB_DEDUP_00057} C. Sa-ngamuang, P. Haddawy, V. Luvira, W. Piyaphanee, S. Iamsirithaworn, S. Lawpoolsri, Accuracy of dengue clinical diagnosis with and without NS1 antigen rapid test: Comparison between human and Bayesian network model decision, PLoS Neglected Tropical Diseases (2018). \url{https://doi.org/10.1371/journal.pntd.0006573}.
\bibitem{SB_DEDUP_00120} J. Quinteros, N. Baloian, J. A. Pino, A. Riquelme, S. Peñafiel, H. Sanson, D. Teoh, Diagnostic test suggestion via Bayesian network of non-expert assisted knowledge base, 2018 20th International Conference on Advanced Communication Technology (2018). \url{https://doi.org/10.23919/ICACT.2018.8323748}.
\bibitem{SB_DEDUP_00105} X. Xiang, Z. Wang, Y. Jia, B. Fang, Knowledge Graph-Based Clinical Decision Support System Reasoning: A Survey, 2019 IEEE 4th International Conference on Data Science in Cyberspace (2019). \url{https://doi.org/10.1109/DSC.2019.00063}.
\bibitem{SB_DEDUP_00169} M. Kaul, N. S. Narayan, A. Narayanan, Siamese bayesian networks for AI based differential diagnosis, Proceedings of the 3rd International Conference on High Performance Compilation, Computing and Communications (2019). \url{https://doi.org/10.1145/3318265.3318298}.
\bibitem{DEDUP_00249} Y. Shen, Y. Li, H.-T. Zheng, B. Tang, M. Yang, Enhancing ontology-driven diagnostic reasoning with a symptom-dependency-aware Naive Bayes classifier, BMC Bioinformatics (2019) 330. \url{https://doi.org/10.1186/s12859-019-2924-0}.
\bibitem{DEDUP_01201} Y. Jia, Z. Tan, J. Zhang, DKDR: An Approach of Knowledge Graph and Deep Reinforcement Learning for Disease Diagnosis, IEEE ISPA/BDCloud/SocialCom/SustainCom (2019) 1303--1308. \url{https://doi.org/10.1109/ISPA-BDCloud-SustainCom-SocialCom48970.2019.00187}.
\bibitem{SB_DEDUP_00528} M. M. Ershadi, A. Seifi, An efficient Bayesian network for differential diagnosis using experts' knowledge, International Journal of Intelligent Computing and Cybernetics (2020). \url{https://doi.org/10.1108/IJICC-10-2019-0112}.
\bibitem{DEDUP_00180} N. Heilig, J. Kirchhoff, F. Stumpe, J. Plepi, L. Flek, H. Paulheim, Refining Diagnosis Paths for Medical Diagnosis based on an Augmented Knowledge Graph, CEUR Workshop Proceedings (2022).
\bibitem{DEDUP_00323} A. Lacki, D. Bosca, A. Martinez-Millana, Probabilistic Inference of Comorbidities from Symptoms in Patients with Atrial Fibrillation: An Ontology-Driven Hybrid Clinical Decision Support System, 2022 Computing in Cardiology (2022) 1--4. \url{https://doi.org/10.22489/CinC.2022.002}.
\bibitem{SB_DEDUP_00479} D. Zhang, Q. Jia, S. Yang, X. Han, C. Xu, X. Li, Y. Xie, Traditional Chinese Medicine Automated Diagnosis Based on Knowledge Graph Reasoning, Computers, Materials and Continua (2022). \url{https://doi.org/10.32604/cmc.2022.017295}.
\bibitem{DEDUP_01193} S. Guo, K. Liu, P. Wang, W. Dai, Y. Du, Y. Zhou, W. Cui, RDKG: A Reinforcement Learning Framework for Disease Diagnosis on Knowledge Graph, 2023 IEEE International Conference on Data Mining (2023) 1049--1054. \url{https://doi.org/10.1109/ICDM58522.2023.00122}.
\bibitem{DEDUP_00212} Y. Shang, Y. Tian, K. Lyu, T. Zhou, P. Zhang, J. Chen, J. Li, Electronic Health Record-Oriented Knowledge Graph System for Collaborative Clinical Decision Support Using Multicenter Fragmented Medical Data: Design and Application Study, Journal of Medical Internet Research (2024) e54263. \url{https://doi.org/10.2196/54263}.
\bibitem{DEDUP_01170} J. M. R. Mejia, D. B. Rawat, ClinicalGraph: An Applied Approach in Clinical EHR Knowledge Graph Generation for Optimized Clinical Decision Support System, 2024 IEEE International Conference on E-health Networking, Application and Services (2024) 1--6. \url{https://doi.org/10.1109/HealthCom60970.2024.10880799}.
\bibitem{DEDUP_02281} Y. Liu, F. Wang, X. Wang, Y. Guo, J. Chang, A Knowledge Graph-Based AI Diagnostic and Reasoning System for Sleep Disorders, Proceedings of the 2025 2nd International Conference on Big Data and Digital Management (2025) 793--797. \url{https://doi.org/10.1145/3768801.3768930}.
\bibitem{DEDUP_01189} J. Du, D. Zhang, F. Luo, H. Su, Research on Automatic Construction Method of Uncertain Knowledge Graph Based on Personal Electronic Medical Records, 2025 10th International Conference on Intelligent Computing and Signal Processing (2025). \url{https://doi.org/10.1109/ICSP65755.2025.11086945}.
\bibitem{DEDUP_00290} J. Zhang, G. Zheng, H. Lv, L. Luo, G. Ma, Z. Lin, X. Chen, Y. Tan, D$^{2}$KGMed: Dynamic Diagnostic Knowledge Graphs for Medical Diagnosis Prediction, 2025 IEEE International Conference on Bioinformatics and Biomedicine (2025) 4458--4461. \url{https://doi.org/10.1109/BIBM66473.2025.11357158}.
\bibitem{SC_03} Y. Gao, R. Li, E. Croxford, J. Caskey, B. W. Patterson, M. Churpek, T. Miller, D. Dligach, M. Afshar, Leveraging Medical Knowledge Graphs Into Large Language Models for Diagnosis Prediction: Design and Application Study, JMIR AI (2025) e58670. \url{https://doi.org/10.2196/58670}.
\bibitem{SB_DEDUP_00569} G. Sowerby, O. Ashaolu, R. Calinescu, S. Connor, Synthesising Bayesian Network Models for Clinical Decision Support from Rule-Based Logic, CEUR Workshop Proceedings (2025).
\bibitem{DEDUP_00001} R. Chavda, K. Suresh, S. Kumar, K. L. R. Reddy, B. Jayaprakash, P. K. Sahu, B. Bharathi, D. Singh, S. Namdev, A bidirectional neuro-symbolic framework for clinical decision support via dynamic integration of deep learning and symbolic reasoning, Network Modeling Analysis in Health Informatics and Bioinformatics (2026) 79. \url{https://doi.org/10.1007/s13721-025-00710-2}.
\bibitem{DEDUP_00227} R. A. Yadav, M. Vaishnavi, J. A. Kurumidde, B. Yalamanchili, Integrating Deep Learning with Symbolic Reasoning: A NeuroSymbolic Framework for Trustworthy Medical Diagnosis, 2026 IEEE International Conference on Emerging Computing and Intelligent Technologies (2026) 1--6. \url{https://doi.org/10.1109/ICoECIT68303.2026.11497919}.
\bibitem{DEDUP_00010} S. Albagli-Kim, D. Beimel, A Semantic Knowledge Graph Approach with Weighted and Conditional Edges for Clinical Reasoning, IFMBE Proceedings (2026) 234--243. \url{https://doi.org/10.1007/978-3-032-24724-7_24}.
\bibitem{DEDUP_03285} J. Dave, P. Patel, I. S. Hakkinen, P. Zhang, A knowledge graph based differential diagnosis engine for headache disorders, Headache (2026). \url{https://doi.org/10.1111/head.70117}.
\bibitem{DEDUP_00352} F. Roucoux, S. Jodogne, Clinical Diagnosis of Rare Diseases Using Leaky Noisy-OR Bayesian Networks, Studies in Health Technology and Informatics (2026) 368--372. \url{https://doi.org/10.3233/SHTI260179}.
\bibitem{DEDUP_00749} N. M. Edward, D. Godwin, F. Odeh, A. Sağlam, M. Leila, A Neuro-Symbolic Expert System for Medical Diagnosis from Natural Language with Rule-Based Logic Engines, 2026 ICHORA (2026). \url{https://doi.org/10.1109/ICHORA69329.2026.11536979}.
\bibitem{DEDUP_01211} Y. Zhao, S. Dong, X. Shi, ReCLLaMA: A Reasoning-Centered LLM Agent for Medical Diagnosis, 2026 IEEE/ACM Conference on Connected Health (2026) 397--402. \url{https://doi.org/10.1109/CHASE69719.2026.00067}.
\bibitem{DEDUP_01192} Y. Chen, X. Zhou, X. Qiao, Y. Lian, W. Zhang, W. Lu, J. Zhu, J. Guo, KGDAgents: A Knowledge Graph Enhanced Multi-agent Framework For Clinical Diagnosis, 2026 29th International Conference on Computer Supported Cooperative Work in Design (2026) 1005--1010. \url{https://doi.org/10.1109/CSCWD68734.2026.11582439}.
\bibitem{SC_01} D. Civale, C. De Maio, D. Furno, S. Senatore, Constructing a clinical knowledge graph from electronic health records for enhanced decision-making and disease diagnosis, Neurocomputing (2026). \url{https://doi.org/10.1016/j.neucom.2025.132358}.
\bibitem{DEDUP_00185} M. He, J. Song, S. Ren, Y. Zhang, J. Du, J. Feng, R. Wu, B. Shen, FPGDKG 1.0: An Integrated Facial Phenotype-Gene-Disease Knowledge Graph for Rare Disease Diagnosis and Explanation, IEEE Journal of Biomedical and Health Informatics (2026) 1--10. \url{https://doi.org/10.1109/JBHI.2026.3659898}.
\bibitem{DEDUP_00114} S. Rancati, L. Bergomi, E. Parimbelli, G. Nicora, R. Bellazzi, Epistemologically Guided LLM Reasoning for Differential Diagnosis, Lecture Notes in Artificial Intelligence 16748: Artificial Intelligence in Medicine (2027) 115--124. \url{https://doi.org/10.1007/978-3-032-30710-1_14}.
\bibitem{SB_DEDUP_00404} J. J. Gonzalez-Lopez, A. M. Garcia-Aparicio, D. Sanchez-Ponce, N. Munoz-Sanz, N. Fernandez-Ledo, P. Beneyto, M. C. Westcott, Development and validation of a Bayesian network for the differential diagnosis of anterior uveitis, Eye (2016). \url{https://doi.org/10.1038/eye.2016.64}.
\end{thebibliography}
\end{document}